\documentclass[runningheads]{llncs}
\usepackage[T1]{fontenc}
\usepackage{graphicx}
\usepackage{booktabs}
\usepackage[misc]{ifsym}

\usepackage[utf8]{inputenc}
\usepackage{amsmath,amssymb,amsfonts}
\usepackage{multirow}
\usepackage{microtype}
\usepackage{adjustbox}
\usepackage{hyperref}
\usepackage{xcolor}

\usepackage{tikz}
\usepackage{pgfplots}
\pgfplotsset{compat=1.18}
\usetikzlibrary{arrows.meta,positioning,calc,shapes.misc,fit,patterns}
\usepgfplotslibrary{groupplots}

\definecolor{axonblue}{HTML}{2171B5}
\definecolor{anomred}{HTML}{CB181D}
\definecolor{nomgreen}{HTML}{238B45}
\definecolor{pastgray}{HTML}{969696}
\definecolor{accentorange}{HTML}{D95F0E}
\definecolor{lightblue}{HTML}{C6DBEF}
\definecolor{lightred}{HTML}{FCBBA1}
\definecolor{lightgreen}{HTML}{C7E9C0}

\newcommand{\myParagraph}[1]{\textbf{#1.}\quad}
\newcommand{\stopgrad}{\operatorname{stopgrad}}
\newcommand{\IQR}{\operatorname{IQR}}
\newcommand{\med}{\operatorname{median}}

\graphicspath{{.}{./figs/}}

\hypersetup{
    hidelinks,
    pdftitle={AxonAD},
}

\begin{document}

\title{Surprised by Attention: Predictable Query Dynamics for Time Series Anomaly Detection}

\titlerunning{Predictable Query Dynamics for Time Series Anomaly Detection}

\author{Kadir-Kaan \"Ozer~(\Letter)\inst{1,2} \and
Ren\'{e} Ebeling\inst{1} \and
Markus Enzweiler\inst{2}}

\authorrunning{K.-K. \"Ozer et al.} 

\institute{Mercedes-Benz AG, Germany\\
\email{\{kadir.oezer, rene.ebeling\}@mercedes-benz.com} \and
Institute for Intelligent Systems, Esslingen University of Applied Sciences, Germany\\
\email{markus.enzweiler@hs-esslingen.de}}

\toctitle{Surprised by Attention: Predictable Query Dynamics for Time Series Anomaly Detection}
\tocauthor{Kadir-Kaan \"Ozer, Ren\'e Ebeling, Markus Enzweiler}

\maketitle

\renewcommand{\thefootnote}{}
\footnotetext{The final publication is available at Springer Nature: \url{https://doi.org/10.1007/978-3-032-37685-5_13}.}
\renewcommand{\thefootnote}{\arabic{footnote}}

\begin{abstract}
Multivariate time series anomalies often manifest as shifts in cross-channel dependencies rather than simple amplitude excursions. In autonomous driving, for instance, a steering command might be internally consistent but decouple from the resulting lateral acceleration. Residual-based detectors can miss such anomalies when flexible sequence models still reconstruct signals plausibly despite altered coordination.

We introduce \textbf{AxonAD}, an unsupervised detector that treats multi-head attention query evolution as a short horizon predictable process. A gradient-updated reconstruction pathway is coupled with a history-only predictor that forecasts future query vectors from past context. This is trained via a masked predictor-target objective against an exponential moving average (EMA) target encoder. At inference, reconstruction error is combined with a tail-aggregated \emph{query mismatch} score, which measures cosine deviation between predicted and target queries on recent timesteps. This dual approach provides sensitivity to structural dependency shifts while retaining amplitude-level detection. On proprietary in-vehicle telemetry with interval annotations and on the TSB-AD multivariate suite (17 datasets, 180 series) with threshold-free and range-aware metrics, AxonAD improves ranking quality and temporal localization over strong baselines. Ablations confirm that query prediction and combined scoring are the primary drivers of the observed gains. Code is available at the URL \url{https://github.com/iis-esslingen/AxonAD}.

\end{abstract}

\section{Introduction}
\label{sec:intro}

Modern vehicles produce dense telemetry streams where different channels, from steering angle and throttle position to lateral acceleration and yaw rate, are sampled at high frequency. Faults in these systems rarely present as individual channels leaving their nominal range. Instead, the typical failure mode is a \emph{coordination break}: a steering command that no longer produces the expected lateral response, or a throttle position that decouples from engine torque. Detecting such anomalies matters directly for fleet monitoring, warranty analytics, and safety validation.

This setting exposes a fundamental limitation of residual-based unsupervised detectors. A flexible sequence model can accurately reconstruct each channel while missing that the joint coordination pattern across channels has changed~\cite{lim_temporal_2021,su_robust_2019,wu_timesnet_2023}. Low reconstruction error does not guarantee that learned representations preserve the full dependency structure.

\begin{figure}[t]
\centering
\begin{tikzpicture}[
  >=Stealth,
  qdot/.style={circle, fill=#1, inner sep=0pt, minimum size=4pt},
  qdot/.default={pastgray},
  predicted marker/.style={circle, draw=axonblue, thick, inner sep=0pt, minimum size=6pt},
  past arrow/.style={->, thick, pastgray},
  target arrow/.style={->, thick, anomred},
  pred arrow/.style={->, thick, densely dashed, axonblue},
]

\begin{scope}
  \node[font=\footnotesize\bfseries, anchor=south] at (1.8, 3.15) {(a) Nominal dynamics};

  \draw[->, very thin, black!15] (-0.2, -0.1) -- (3.8, -0.1);
  \draw[->, very thin, black!15] (-0.1, -0.2) -- (-0.1, 2.8);

  \node[qdot] (a0) at (0.3, 0.4) {};
  \node[qdot] (a1) at (1.1, 0.9) {};
  \node[qdot] (a2) at (2.0, 1.4) {};

  \node[qdot=anomred, label={[font=\tiny, anomred, xshift=3pt]right:{$\mathbf{q}^{\mathrm{tgt}}_{\tau}$}}] (atgt) at (2.85, 1.85) {};

  \node[predicted marker, label={[font=\tiny, axonblue, xshift=-3pt]above left:{$\hat{\mathbf{q}}^{\mathrm{pred}}_{\tau}$}}] (apred) at (2.78, 1.95) {};

  \draw[past arrow] (a0) -- (a1);
  \draw[past arrow] (a1) -- (a2);

  \draw[target arrow] (a2) -- (atgt);
  \draw[pred arrow] (a2) -- (apred);

  \node[font=\tiny, pastgray, below left, inner sep=1pt] at (a0) {$\tau{-}3$};
  \node[font=\tiny, pastgray, below, inner sep=1pt] at (a1) {$\tau{-}2$};
  \node[font=\tiny, pastgray, below left, inner sep=1pt] at (a2) {$\tau{-}1$};

  \node[font=\scriptsize, nomgreen, align=center] at (1.8, -0.55) {$d_q \approx 0$: match};
\end{scope}

\begin{scope}[xshift=5.8cm]
  \node[font=\footnotesize\bfseries, anchor=south] at (1.8, 3.15) {(b) Coordination anomaly};

  \draw[->, very thin, black!15] (-0.2, -0.1) -- (3.8, -0.1);
  \draw[->, very thin, black!15] (-0.1, -0.2) -- (-0.1, 2.8);

  \node[qdot] (b0) at (0.3, 0.4) {};
  \node[qdot] (b1) at (1.1, 0.9) {};
  \node[qdot] (b2) at (2.0, 1.4) {};

  \node[predicted marker, label={[font=\tiny, axonblue, xshift=3pt]right:{$\hat{\mathbf{q}}^{\mathrm{pred}}_{\tau}$}}] (bpred) at (2.85, 1.85) {};

  \node[qdot=anomred, label={[font=\tiny, anomred, yshift=2pt]above:{$\mathbf{q}^{\mathrm{tgt}}_{\tau}$}}] (btgt) at (2.35, 2.7) {};

  \draw[past arrow] (b0) -- (b1);
  \draw[past arrow] (b1) -- (b2);

  \draw[pred arrow] (b2) -- (bpred);

  \draw[target arrow] (b2) -- (btgt);

  \draw[thin, anomred!60] ($(b2)+(24:0.8)$) arc[start angle=24, end angle=69, radius=0.8];

  \node[font=\tiny, pastgray, below left, inner sep=1pt] at (b0) {$\tau{-}3$};
  \node[font=\tiny, pastgray, below, inner sep=1pt] at (b1) {$\tau{-}2$};
  \node[font=\tiny, pastgray, below left, inner sep=1pt] at (b2) {$\tau{-}1$};

  \node[font=\scriptsize, anomred, align=center] at (1.8, -0.55) {$d_q \gg 0$: mismatch};
\end{scope}

\end{tikzpicture}
\caption{Query predictability in 2D query space (schematic, single head). \textcolor{pastgray}{Gray}: past query trajectory. \textcolor{axonblue}{Blue dashed}: predicted query. \textcolor{anomred}{Red}: EMA target query. (a)~Nominal: predictor and target agree. (b)~Coordination anomaly: the target query diverges from the predicted trajectory, producing large $d_q$ even when per-channel amplitudes are within normal bounds.}
\label{fig:query_concept}
\end{figure}
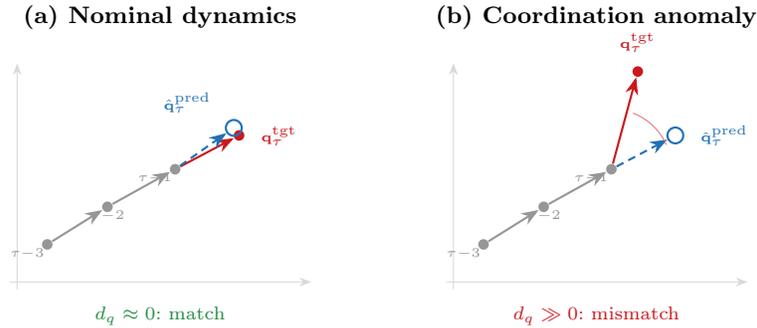

Attention mechanisms~\cite{vaswani2017attentionneed} capture relational structure through query and key matching, but are typically treated as a one-shot computation for the current input window. Under stationary nominal dynamics, the query vectors that control attention routing should evolve predictably over short horizons. Structural anomalies can disrupt this predictability even when per-channel amplitudes remain plausible, making query mismatch a diagnostic signal complementary to reconstruction error. Figure~\ref{fig:query_concept} illustrates this idea.

AxonAD combines two coupled pathways. The first reconstructs the input window using bidirectional self attention. The second is a history-only predictor that maps a time-shifted embedding stream to future multi-head query vectors, trained with a masked cosine loss against an exponential moving average (EMA) target encoder. At inference, reconstruction error and query mismatch are each robustly standardized on nominal training data and summed to produce the final anomaly score.

We evaluate on proprietary in-vehicle telemetry with interval annotations as the primary setting and on the multivariate TSB-AD suite~\cite{paparrizos_volume_2022,liu_elephant_2024}. Across both, AxonAD improves threshold-free ranking and temporal localization relative to strong baselines, and ablations confirm that query prediction and score combination are the primary drivers.

\noindent Our contributions are:
\begin{itemize}
\item A predictive attention anomaly detector that treats query vectors as a temporally predictable signal rather than a one-shot routing decision, providing sensitivity to structural dependency shifts.
\item Query mismatch as a tail-focused anomaly score that complements reconstruction residuals with a cosine distance signal in query space.
\item A stable training scheme based on EMA predictor and target networks with masked supervision, avoiding direct supervision on attention maps or value outputs.
\end{itemize}

\section{Related Work}
\label{sec:related}

\subsection{Classical Unsupervised Multivariate Detection}
Isolation-based methods flag anomalies as points that are easily separated under
random partitioning~\cite{hariri_extended_2021,liu_isolation_2008}. Density and
neighborhood methods detect samples whose local geometry differs from the nominal
distribution~\cite{breunig_lof_2000,cover_knn_1967}. Robust matrix decomposition
approaches model data as low-rank structure plus sparse
corruption~\cite{candes_robust_2011}, and clustering, histogram, and copula-based
methods extend this family with alternative density
surrogates~\cite{yairi_kmeans_2001,He_CBLOF_2003,goldstein_hbos_2012,li_copod_2020}.
Because none of these methods capture context-dependent coupling that varies across
operating modes, they have limited sensitivity to coordination-type anomalies.

\subsection{Deep Sequence Models with Residual or Likelihood Scoring}
Deep detectors learn nominal dynamics through reconstruction or forecasting and score
anomalies by residual magnitude or likelihood deviation. Recurrent
reconstruction~\cite{malhotra_lstm_2015} and probabilistic variants such as VAE and
stochastic recurrent models~\cite{li_anomaly_2021,pereira_unsupervised_2018,pereira_unsupervised_2019,su_robust_2019,xu_unsupervised_2018} perform well across
many benchmarks, as do lightweight spectral and convolutional
variants~\cite{xu_fits_2024}. However, residual scoring can miss anomalies that shift
dependencies while leaving per-channel values plausible, particularly under
nonstationarity where flexible models may still reconstruct accurately despite altered
coordination~\cite{lim_temporal_2021,su_robust_2019,wu_timesnet_2023}.

\subsection{Attention and Relation Aware Anomaly Scoring}
Attention weights encode learned relational structure~\cite{vaswani2017attentionneed}
and have been used directly for scoring, for example by measuring association
discrepancies~\cite{xu_anomaly_2022} or by modeling sensor relations with graph
structures~\cite{deng_graph_2021}. Transformer backbones have also been adapted to
anomaly detection through reconstruction and forecasting
pipelines~\cite{lim_temporal_2021,tuli_tranad_2022,wu_timesnet_2023,zhou2023fitsallpowergeneraltime}. AxonAD differs in that it scores the
\emph{predictability} of query vectors over time, capturing what the model is about
to attend to, rather than scoring the attention weights themselves or the value
residuals.

\subsection{Self-Supervised Predictive Objectives}
Predictive self-supervised learning encourages representations to be inferable from context under masking, commonly stabilized via EMA target networks~\cite{assran_self-supervised_2023,balestriero_lejepa_2025}. Related masking objectives have been applied to time series representation learning~\cite{abrantes_competition_2025,xu_fits_2024,zhou2023fitsallpowergeneraltime}. Most detectors that use prediction supervise values or latent states and score residuals at inference. AxonAD instead applies predictive supervision directly in query space, making the training objective and the inference scoring signal the same cosine distance. Section~\ref{subsec:score} exploits this consistency.

\section{Model Architecture}
\label{sec:arch}

\begin{figure}[!t]
  \centering
  \includegraphics[width=1.0\textwidth]{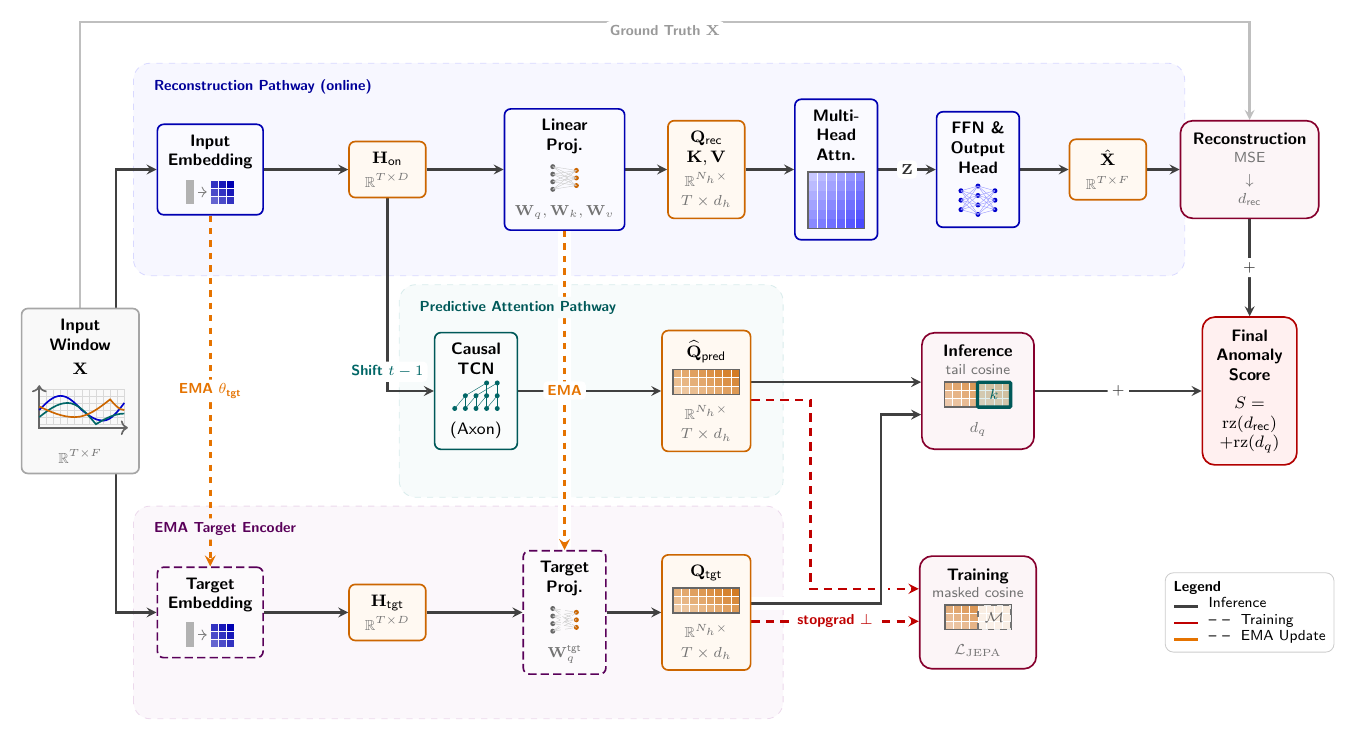}
  \caption{AxonAD overview. The online reconstruction encoder computes self attention using queries
  $\mathbf{Q}_{\mathrm{rec}}$. In parallel, a history-only predictor forecasts
  $\widehat{\mathbf{Q}}_{\mathrm{pred}}$ and is trained to match EMA target
  queries $\mathbf{Q}_{\mathrm{tgt}}$ (stop-gradient). Query mismatch on the
  last valid timesteps yields $d_q$, and reconstruction yields $d_{\mathrm{rec}}$.
  Attention divergence (KL tail) is not included in the default scoring pipeline.}
  \label{fig:axon_arch}
\end{figure}

Figure~\ref{fig:axon_arch} gives an overview. The model takes a fixed-length window $\mathbf{X}\in\mathbb{R}^{T\times F}$ and produces a reconstruction $\hat{\mathbf{X}}\in\mathbb{R}^{T\times F}$ together with two window-level signals: a reconstruction score $d_{\mathrm{rec}}$ and a query mismatch score $d_q$, combined after robust standardization into the final anomaly score.

The architecture has three components: (i) a gradient-updated reconstruction pathway based on bidirectional self attention, (ii) a history-only predictive pathway that forecasts future multi-head query vectors from a time-shifted embedding stream, and (iii) an EMA target encoder that provides stable query supervision targets~\cite{grill2020bootstraplatentnewapproach}. Throughout this paper, \emph{online} refers to gradient-updated parameters, not streaming causality.

\myParagraph{Notation}
$T$ denotes the window length, $F$ the number of channels, and $D$ the embedding dimension. $N_h$ is the number of attention heads with head dimension $d_h=D/N_h$. $s$ is the forecast horizon, $k$ the number of tail timesteps used for query mismatch aggregation, and $m\in(0,1)$ is the EMA momentum. We use $\tau\in\{1,\dots,T\}$ for within window timestep indices. For a window ending at absolute time $t$, we write $\mathbf{X}_{t-T+1:t}\in\mathbb{R}^{T\times F}$ for its $T$ rows.

\myParagraph{Shared embedding}
A linear projection with learnable positional bias maps $\mathbf{X}$ to a shared per timestep representation, followed by layer normalization~\cite{ba2016layernormalization} applied before attention:
\[
\mathbf{H}_{\mathrm{on}} = \mathrm{LN}\!\left(\mathbf{X}\mathbf{W}_e + \mathbf{b}_e + \mathbf{P}\right),
\qquad \mathbf{P}\in\mathbb{R}^{T\times D}.
\]
This sequence feeds both the reconstruction self attention and the predictive branch (after applying the history-only time shift described below).

\subsection{Online Reconstruction Pathway}
\label{subsec:rec_path}

The online encoder forms multi-head queries, keys, and values via learned projections:
\[
(\mathbf{Q}_{\mathrm{rec}},\mathbf{K},\mathbf{V}) \in
\mathbb{R}^{N_h\times T\times d_h}.
\]
Standard multi-head self attention~\cite{vaswani2017attentionneed} over full within window context produces context features $\mathbf{Z}$, which are processed by a position-wise feedforward network and a linear output head to obtain $\hat{\mathbf{X}}$. The reconstruction score is the mean squared $\ell_2$ error over timesteps:
\begin{equation}
\label{eq:drec}
d_{\mathrm{rec}}
=
\frac{1}{T}\sum_{\tau=1}^{T}
\left\lVert \hat{\mathbf{x}}_{\tau}-\mathbf{x}_{\tau}\right\rVert_{2}^{2}.
\end{equation}

\subsection{Predictive Attention Pathway}
\label{subsec:pred_path}

The predictive pathway forecasts query vector evolution using only past context, producing an anomaly signal sensitive to coordination shifts even when windows remain plausible in amplitude.

\myParagraph{History-only shift}
To prevent information leakage, we construct a time-shifted embedding stream with forecast horizon $s$:
\[
\widetilde{\mathbf{H}}_{\tau}=
\begin{cases}
\mathbf{0}, & \tau\le s,\\
\mathbf{H}_{\mathrm{on},\,\tau-s}, & \tau>s,
\end{cases}
\]
ensuring that any prediction at timestep $\tau$ depends only on embeddings available up to $\tau-s$.

\myParagraph{Causal predictor}
A causal temporal predictor $g(\cdot)$ maps the shifted sequence to predicted multi-head queries:
\[
\widehat{\mathbf{Q}}_{\mathrm{pred}} = g(\widetilde{\mathbf{H}})
\in \mathbb{R}^{N_h\times T\times d_h},
\]
with causality enforced so that the output at $\tau$ depends only on $\widetilde{\mathbf{H}}_{\le \tau}$. We denote the per head, per timestep slice by $\widehat{\mathbf{q}}^{\mathrm{pred}}_{h,\tau} = \widehat{\mathbf{Q}}_{\mathrm{pred}}[h,\tau,:]\in\mathbb{R}^{d_h}$, with the corresponding EMA target slice $\mathbf{q}^{\mathrm{tgt}}_{h,\tau}$ defined in Section~\ref{subsec:ema_train}. The predictive branch forecasts queries only, not keys or values, keeping it lightweight and aligning supervision directly with the inference scoring signal.

\subsection{EMA Target Encoder and Masked Training}
\label{subsec:ema_train}

We maintain an EMA target encoder with parameters $\theta_{\mathrm{tgt}}$ that track the online parameters $\theta_{\mathrm{on}}$:
\[
\theta_{\mathrm{tgt}}\leftarrow m\,\theta_{\mathrm{tgt}}+(1-m)\,\theta_{\mathrm{on}},
\qquad m\in(0,1),
\]
with no gradient updates to the target parameters~\cite{grill2020bootstraplatentnewapproach}. Given the same input window $\mathbf{X}$, the EMA encoder produces a target embedding sequence $\mathbf{H}_{\mathrm{tgt}}\in\mathbb{R}^{T\times D}$ in the same way as $\mathbf{H}_{\mathrm{on}}$ but using $\theta_{\mathrm{tgt}}$. Target queries are obtained via a mirrored projection:
\begin{equation}
\label{eq:qtgt}
\mathbf{Q}_{\mathrm{tgt}}
=
\mathrm{reshape}_{N_h}\!\left(\mathbf{H}_{\mathrm{tgt}}\mathbf{W}^{\mathrm{tgt}}_{q}\right)
\in\mathbb{R}^{N_h\times T\times d_h},
\end{equation}
where $\mathbf{W}^{\mathrm{tgt}}_{q}$ is the EMA tracked counterpart of the online query projection.

Training minimizes reconstruction error together with a masked cosine loss in query space, following a JEPA style scheme~\cite{assran_self-supervised_2023}. A set of masked timesteps $\mathcal{M}\subset\{s+1,\dots,T\}$ is sampled via contiguous time patch masking over valid timesteps (inputs remain unmasked). The resulting loss is:
\begin{equation}
\label{eq:Ljepa}
\mathcal{L}_{\mathrm{JEPA}}
=
\frac{1}{|\mathcal{M}|\,N_h}
\sum_{\tau\in \mathcal{M}}
\sum_{h=1}^{N_h}
\Bigl(
1 -
\Bigl\langle
\frac{\widehat{\mathbf{q}}^{\mathrm{pred}}_{h,\tau}}{\|\widehat{\mathbf{q}}^{\mathrm{pred}}_{h,\tau}\|_2+\varepsilon_{\cos}},\,
\frac{\stopgrad(\mathbf{q}^{\mathrm{tgt}}_{h,\tau})}{\|\stopgrad(\mathbf{q}^{\mathrm{tgt}}_{h,\tau})\|_2+\varepsilon_{\cos}}
\Bigr\rangle
\Bigr).
\end{equation}
The stop-gradient on $\mathbf{q}^{\mathrm{tgt}}_{h,\tau}$ ensures that only the predictor is updated to match the targets, not the reverse.

\subsection{Query Mismatch and Final Anomaly Score}
\label{subsec:score}

At inference, AxonAD computes two complementary window-level signals: $d_{\mathrm{rec}}$ (Eq.~\eqref{eq:drec}) and a query mismatch score $d_q$ derived from cosine deviations between predicted and EMA target queries on the \emph{tail} of the window, emphasizing the most recent timesteps.

The tail-aggregated query mismatch is defined as:
\begin{equation}
\label{eq:dq}
\begin{aligned}
\tau_0
&=
\max\!\bigl(s+1,\,T-k+1\bigr),\\
k_{\mathrm{eff}}
&=
T-\tau_0+1,\\
d_q
&=
\frac{1}{N_h \, k_{\mathrm{eff}}}
\sum_{h=1}^{N_h}
\sum_{\tau=\tau_0}^{T}
\Bigl(
1 -
\Bigl\langle
\frac{\widehat{\mathbf{q}}^{\mathrm{pred}}_{h,\tau}}{\|\widehat{\mathbf{q}}^{\mathrm{pred}}_{h,\tau}\|_2+\varepsilon_{\cos}},
\frac{\mathbf{q}^{\mathrm{tgt}}_{h,\tau}}{\|\mathbf{q}^{\mathrm{tgt}}_{h,\tau}\|_2+\varepsilon_{\cos}}
\Bigr\rangle
\Bigr),
\end{aligned}
\end{equation}
where $\tau_0$ enforces both validity under the $s$ step history constraint and tail focus of nominal length $k$, and $k_{\mathrm{eff}}$ normalizes by the actual number of summed timesteps.

Because $d_{\mathrm{rec}}$ and $d_q$ can have very different dynamic ranges across datasets, each component is robustly standardized using median and interquartile range (IQR) statistics computed exclusively on nominal training windows:
\begin{equation}
\label{eq:robustz}
\mathrm{rz}(u)=\frac{u-\med(u)}{\IQR(u)+\varepsilon_{\mathrm{rz}}},
\qquad
\IQR(u)=Q_{0.75}(u)-Q_{0.25}(u),
\end{equation}
and the final anomaly score is:
\begin{equation}
\label{eq:score}
S(\mathbf{X})
=
\mathrm{rz}\!\left(d_{\mathrm{rec}}(\mathbf{X})\right)
+
\mathrm{rz}\!\left(d_q(\mathbf{X})\right).
\end{equation}

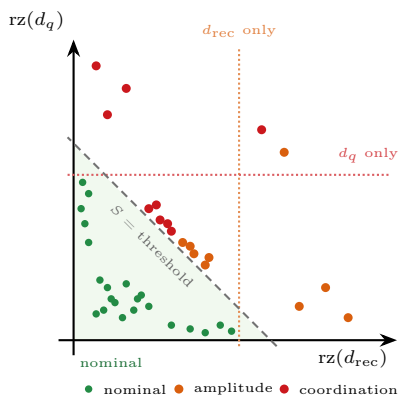
\begin{figure}[t]
\centering
\begin{tikzpicture}[>=Stealth]

  \fill[lightgreen, opacity=0.25]
    (0.0, 0.0) -- (2.6, 0.0) -- (0.0, 2.6) -- cycle;

  \draw[->, thick] (-0.2, 0) -- (4.3, 0)
    node[below left, font=\scriptsize] {$\mathrm{rz}(d_{\mathrm{rec}})$};
  \draw[->, thick] (0, -0.2) -- (0, 4.0)
    node[above left, font=\scriptsize, align=right] {$\mathrm{rz}(d_q)$};

  \draw[densely dotted, thick, accentorange!70] (2.2, -0.08) -- (2.2, 3.9);
  \node[font=\tiny, accentorange!80, anchor=south] at (2.2, 3.92)
    {$d_{\mathrm{rec}}$ only};

  \draw[densely dotted, thick, anomred!70] (-0.08, 2.2) -- (4.2, 2.2);
  \node[font=\tiny, anomred!80, anchor=south west] at (3.4, 2.25)
    {$d_q$ only};

  \draw[thick, densely dashed, black!55] (2.7, -0.08) -- (-0.08, 2.7);
  \node[font=\tiny, black!60, rotate=-45, anchor=north east, yshift=-1pt]
    at (1.8, 0.8) {$S$ = threshold};


  \foreach \x/\y in {
    0.30/0.35, 0.50/0.55, 0.80/0.40, 0.45/0.70,
    0.85/0.55, 0.65/0.30, 1.00/0.45, 0.35/0.80,
    0.90/0.60, 0.70/0.75, 0.55/0.50, 0.40/0.40} {
    \fill[nomgreen] (\x, \y) circle (1.4pt);
  }
  \foreach \x/\y in {
    1.30/0.20, 1.55/0.15, 1.75/0.10, 1.95/0.20, 2.10/0.12} {
    \fill[nomgreen] (\x, \y) circle (1.4pt);
  }
  \foreach \x/\y in {
    0.20/1.30, 0.15/1.55, 0.10/1.75, 0.20/1.95, 0.12/2.10} {
    \fill[nomgreen] (\x, \y) circle (1.4pt);
  }

  \foreach \x/\y in {
    1.45/1.30, 1.60/1.15, 1.75/1.00, 1.55/1.25, 1.80/1.10} {
    \fill[accentorange] (\x, \y) circle (1.7pt);
  }

  \foreach \x/\y in {
    1.30/1.45, 1.15/1.60, 1.00/1.75, 1.25/1.55, 1.10/1.80} {
    \fill[anomred] (\x, \y) circle (1.7pt);
  }

  \foreach \x/\y in {3.00/0.45, 3.35/0.70, 3.65/0.30} {
    \fill[accentorange] (\x, \y) circle (1.7pt);
  }

  \foreach \x/\y in {0.45/3.00, 0.70/3.35, 0.30/3.65} {
    \fill[anomred] (\x, \y) circle (1.7pt);
  }

  \fill[accentorange] (2.80, 2.50) circle (1.7pt);
  \fill[anomred] (2.50, 2.80) circle (1.7pt);

  \node[font=\tiny, nomgreen!80!black] at (0.50, -0.30) {nominal};

  \fill[nomgreen] (0.20, -0.65) circle (1.4pt);
  \node[font=\tiny, right, inner sep=2pt] at (0.32, -0.65) {nominal};
  \fill[accentorange] (1.40, -0.65) circle (1.7pt);
  \node[font=\tiny, right, inner sep=2pt] at (1.52, -0.65) {amplitude};
  \fill[anomred] (2.80, -0.65) circle (1.7pt);
  \node[font=\tiny, right, inner sep=2pt] at (2.92, -0.65) {coordination};

\end{tikzpicture}
\caption{Score complementarity (schematic).
  \textcolor{nomgreen!80!black}{Nominal} windows spread along both axes but cluster near the origin in both scores simultaneously. Near-boundary \textcolor{accentorange}{amplitude} and \textcolor{anomred}{coordination} anomalies are moderate on both axes, falling inside both single-axis thresholds (dotted lines) but separated by the additive $S$ (dashed diagonal).}
\label{fig:score_landscape}
\end{figure}

\noindent The additive form means that a single threshold on $S$ captures anomalies that elevate either component or both. Figure~\ref{fig:score_landscape} illustrates the geometry: amplitude anomalies raise $d_{\mathrm{rec}}$ while coordination anomalies raise $d_q$, and the diagonal constant score contour separates all anomaly types from the nominal cluster.

\myParagraph{Training and inference consistency}
The cosine distance used for masked supervision in Eq.~\eqref{eq:Ljepa} is the same metric reused at inference as $d_q$ in Eq.~\eqref{eq:dq}. This means the predictor is trained directly on the deployed scoring objective. An attention divergence diagnostic (KL tail) is implemented for ablation analysis only and is not part of the default scoring pipeline.

\section{Experimental Setup}
\label{sec:exp}

\myParagraph{Protocol}
We evaluate in two settings: (i) proprietary in-vehicle telemetry with interval annotations, and (ii) the TSB-AD multivariate suite (17 datasets, 180 series) under the official pipeline~\cite{liu_elephant_2024,paparrizos_volume_2022}. Training is strictly unsupervised. All parameters and robust scoring statistics are fit on nominal training windows only, with labels reserved for evaluation. Hyperparameters for all methods are selected on the official TSB-AD tuning component (20 multivariate series) and then fixed. Telemetry labels are never used for hyperparameter selection, thresholding, postprocessing, or early stopping. Early stopping uses a fixed criterion (validation reconstruction error) on an unlabeled split carved from the nominal training prefix.

\myParagraph{Label-free transfer check}
To verify that hyperparameters selected on TSB-AD transfer reasonably to the telemetry domain, we compare distributional similarity using z-scored summary features (scale, shape, autocorrelation, and spectral descriptors) computed on train segments. The telemetry segment is not an outlier: its leave-one-out Mahalanobis distance falls at the 45th percentile and its nearest-neighbor distance at the 55th percentile.

\subsection{Datasets, Splits, and Windowing}

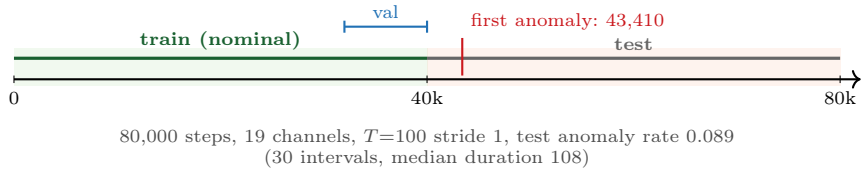
\begin{figure}[t]
\centering
\begin{tikzpicture}[x=0.00013cm,y=1cm]

  \fill[lightgreen, opacity=0.25] (0,-0.08) rectangle (40000,0.42);
  \fill[lightred, opacity=0.18] (40000,-0.08) rectangle (80000,0.42);

  \draw[->,thick] (0,0) -- (82000,0);

  \draw[very thick, nomgreen!70!black] (0,0.28) -- (40000,0.28);
  \node[above, font=\scriptsize\bfseries, nomgreen!70!black] at (20000,0.28) {train (nominal)};

  \draw[very thick, black!60] (40000,0.28) -- (80000,0.28);
  \node[above, font=\scriptsize\bfseries, black!60] at (60000,0.28) {test};

  \draw[thick, axonblue] (32000,0.70) -- (40000,0.70);
  \draw[thick, axonblue] (32000,0.62) -- (32000,0.78);
  \draw[thick, axonblue] (40000,0.62) -- (40000,0.78);
  \node[above, font=\scriptsize, axonblue] at (36000,0.70) {val};

  \draw[thick, anomred] (43410,0.05) -- (43410,0.55);
  \node[above right, font=\scriptsize, anomred] at (43410,0.55) {first anomaly: 43{,}410};

  \foreach \x/\lab in {0/0,40000/40k,80000/80k} {
    \draw (\x,0.05) -- (\x,-0.05);
    \node[below, font=\scriptsize] at (\x,-0.05) {\lab};
  }

  \node[below, font=\scriptsize, black!70, align=center] at (40000,-0.55) {
    80{,}000 steps, 19 channels, $T{=}100$ stride 1, test anomaly rate 0.089 \\
    (30 intervals, median duration 108)
  };

\end{tikzpicture}
\caption{Chronological split and anomaly onset for the proprietary telemetry stream.}
\label{fig:telemetry_split}
\end{figure}

The proprietary telemetry stream contains 80{,}000 timesteps with $F{=}19$ continuous channels (Figure~\ref{fig:telemetry_split}). Anomalies are annotated as contiguous intervals (30 total, duration 1 to 292 with median 108, affecting 1 to 4 channels with median 2) spanning the following types: flatline, drift, level shift, spike, variance jump, and correlation break. The chronological split is: train $[0,40000)$ with an internal 20\% validation holdout (train$\_$sub $[0,32000)$, val $[32000,40000)$), and test $[40000,80000)$. The first anomaly occurs at index 43{,}410, so both training and validation partitions are anomaly free.

The TSB-AD multivariate suite aggregates 180 series across 17 datasets~\cite{paparrizos_volume_2022,liu_elephant_2024}. We follow the official evaluator and protocol throughout.

\myParagraph{Causality and latency}
Window scoring uses no lookahead: $S(\mathbf{X}_{t-T+1:t})$ depends only on samples up to $t$. For real-time deployment, each window score is naturally assigned to its endpoint $t$ (detection time). However, to comply with the point-wise metric computation of the TSB-AD evaluation framework, offline benchmark scores are assigned to the center of the window at $t-\lfloor(T-1)/2\rfloor$. This sequence alignment applies boundary edge padding and effectively incorporates a lookahead of $\lfloor(T-1)/2\rfloor$ steps solely for temporal localization evaluation. Reconstruction attention remains bidirectional within each window, while query prediction is history-only via the $s$ step shift.

\subsection{Baselines and Metrics}
We compare against classical, deep reconstruction and forecasting, and Trans\-former-based detectors implemented in the official TSB-AD framework: Isolation Forest~\cite{liu_isolation_2008}, Extended Isolation Forest~\cite{hariri_extended_2021}, LSTMAD~\cite{malhotra_lstm_2015}, OmniAnomaly~\cite{su_robust_2019}, USAD~\cite{audibert_usad_2020}, VAE variants~\cite{correia_ma-vae_2023,li_anomaly_2021,pereira_unsupervised_2018,pereira_unsupervised_2019} including VASP~\cite{von_schleinitz_vasp_2021}, and Transformer-based baselines (TFTResidual~\cite{lim_temporal_2021}, TimesNet~\cite{wu_timesnet_2023}, TranAD~\cite{tuli_tranad_2022}, Anomaly Transformer~\cite{xu_anomaly_2022}). The main paper reports a representative subset, with full results in the Appendix.

We report threshold-free ranking via AUC-ROC, AUC-PR, VUS-ROC, and VUS-PR~\cite{boniol2025vuseffectiveefficientaccuracy}, and localization via PA-F1, Event-F1, Range-F1, and Affiliation-F1 using the official evaluator~\cite{liu_elephant_2024,paparrizos_volume_2022}. For F1 family metrics, operating points follow the evaluator's default threshold sweep (oracle).

\subsection{AxonAD Configuration}
A single configuration is used across all datasets. The model applies a linear embedding with learnable positional bias ($\mathcal{N}(0,0.02)$), prenorm multi-head self attention, and a feedforward network of width $2D$ with ReLU. The predictive branch is a causal temporal convolutional network~\cite{tcn_2016} with dilations $(1,2,4,8)$, kernel size 3, and dropout 0.1. The EMA target encoder~\cite{grill2020bootstraplatentnewapproach} is initialized from the online model and updated each step with momentum $m{=}0.9$. Query supervision uses time patch masking focused on later timesteps (mask ratio 0.5, block fraction 0.5). Training minimizes reconstruction MSE plus cosine query prediction loss with uncertainty weighting~\cite{cipolla2018uncertaintyweighting}, optimized via AdamW~\cite{loshchilov2019decoupledweightdecayregularization} (weight decay $10^{-5}$), gradient clipping at 1.0, and early stopping on validation reconstruction error.

Unless stated otherwise, reported results use $T{=}100$, $D{=}128$, 8 attention heads, forecast horizon $s{=}1$, tail length $k{=}10$, learning rate $5{\times}10^{-4}$, batch size 128, and up to 50 epochs with patience 3. Results are averaged over four seeds $\{2024,\dots,2027\}$. All experiments have been run on a single Apple MacBook Pro (M3 Max, 32\,GB unified memory) using PyTorch with Apple Silicon acceleration.

\section{Results}
\label{sec:results}

We first report results on the proprietary telemetry stream, which is the primary applied setting, and then on the TSB-AD benchmark to assess generalization.

\begin{table}[t]
\centering
\caption{Proprietary telemetry results (TSB-AD evaluation suite). Mean $\pm$ std over four seeds. Best per metric in \textbf{bold}.}
\label{tab:telemetry}
\resizebox{\textwidth}{!}{%
\begin{tabular}{@{}lcccccccc@{}}
\toprule
\multirow{2}{*}{\textbf{Model}}
& \multicolumn{8}{c}{\textbf{Proprietary Telemetry (19 channels, 80{,}000 timesteps)}} \\ \cmidrule(l){2-9}
& \textbf{AUC-PR} & \textbf{AUC-ROC} & \textbf{VUS-PR} & \textbf{VUS-ROC}
& \textbf{PA-F1} & \textbf{Event-F1} & \textbf{Range-F1} & \textbf{Affiliation-F1} \\
\midrule
LSTMAD~\cite{malhotra_lstm_2015} & 0.082 $\pm$ 0.004 & 0.651 $\pm$ 0.009 & 0.083 $\pm$ 0.004 & 0.624 $\pm$ 0.009 & 0.533 $\pm$ 0.014 & 0.255 $\pm$ 0.015 & 0.139 $\pm$ 0.006 & 0.723 $\pm$ 0.003 \\
SISVAE~\cite{li_anomaly_2021} & 0.128 $\pm$ 0.030 & 0.586 $\pm$ 0.026 & 0.070 $\pm$ 0.012 & 0.504 $\pm$ 0.052 & 0.270 $\pm$ 0.100 & 0.231 $\pm$ 0.060 & 0.225 $\pm$ 0.054 & 0.699 $\pm$ 0.018 \\
TFTResidual~\cite{lim_temporal_2021} & 0.071 $\pm$ 0.006 & 0.644 $\pm$ 0.025 & 0.070 $\pm$ 0.005 & 0.582 $\pm$ 0.018 & 0.424 $\pm$ 0.022 & 0.164 $\pm$ 0.019 & 0.110 $\pm$ 0.009 & \textbf{0.752 $\pm$ 0.026} \\
VSVAE~\cite{pereira_unsupervised_2018} & 0.100 $\pm$ 0.005 & 0.617 $\pm$ 0.031 & 0.065 $\pm$ 0.005 & 0.535 $\pm$ 0.027 & 0.214 $\pm$ 0.048 & 0.188 $\pm$ 0.004 & 0.262 $\pm$ 0.037 & 0.730 $\pm$ 0.012 \\
M2N2~\cite{abrantes_competition_2025} & 0.065 $\pm$ 0.001 & 0.596 $\pm$ 0.001 & 0.064 $\pm$ 0.001 & 0.553 $\pm$ 0.004 & 0.392 $\pm$ 0.022 & 0.196 $\pm$ 0.020 & 0.120 $\pm$ 0.009 & 0.680 $\pm$ 0.000 \\
MAVAE~\cite{correia_ma-vae_2023} & 0.094 $\pm$ 0.006 & 0.561 $\pm$ 0.034 & 0.059 $\pm$ 0.006 & 0.487 $\pm$ 0.051 & 0.220 $\pm$ 0.076 & 0.199 $\pm$ 0.015 & 0.202 $\pm$ 0.017 & 0.680 $\pm$ 0.000 \\
VASP~\cite{von_schleinitz_vasp_2021} & 0.050 $\pm$ 0.001 & 0.540 $\pm$ 0.014 & 0.051 $\pm$ 0.002 & 0.449 $\pm$ 0.016 & 0.190 $\pm$ 0.008 & 0.099 $\pm$ 0.004 & 0.119 $\pm$ 0.013 & 0.686 $\pm$ 0.008 \\
WVAE~\cite{pereira_unsupervised_2019} & 0.087 $\pm$ 0.013 & 0.541 $\pm$ 0.043 & 0.057 $\pm$ 0.007 & 0.467 $\pm$ 0.050 & 0.249 $\pm$ 0.103 & 0.226 $\pm$ 0.038 & 0.163 $\pm$ 0.011 & 0.680 $\pm$ 0.000 \\
TimesNet~\cite{wu_timesnet_2023} & 0.055 $\pm$ 0.001 & 0.579 $\pm$ 0.003 & 0.056 $\pm$ 0.000 & 0.531 $\pm$ 0.003 & 0.306 $\pm$ 0.020 & 0.102 $\pm$ 0.004 & 0.092 $\pm$ 0.002 & 0.680 $\pm$ 0.000 \\
IForest~\cite{liu_isolation_2008} & 0.041 $\pm$ 0.000 & 0.472 $\pm$ 0.000 & 0.044 $\pm$ 0.000 & 0.328 $\pm$ 0.000 & 0.140 $\pm$ 0.000 & 0.086 $\pm$ 0.000 & 0.195 $\pm$ 0.000 & 0.682 $\pm$ 0.000 \\
TranAD~\cite{tuli_tranad_2022} & 0.041 $\pm$ 0.000 & 0.470 $\pm$ 0.003 & 0.044 $\pm$ 0.000 & 0.417 $\pm$ 0.004 & 0.237 $\pm$ 0.003 & 0.086 $\pm$ 0.000 & 0.107 $\pm$ 0.007 & 0.680 $\pm$ 0.000 \\
USAD~\cite{audibert_usad_2020} & 0.040 $\pm$ 0.001 & 0.470 $\pm$ 0.010 & 0.044 $\pm$ 0.001 & 0.371 $\pm$ 0.011 & 0.122 $\pm$ 0.005 & 0.087 $\pm$ 0.001 & 0.152 $\pm$ 0.033 & 0.682 $\pm$ 0.001 \\
OmniAnomaly~\cite{su_robust_2019} & 0.041 $\pm$ 0.000 & 0.459 $\pm$ 0.000 & 0.043 $\pm$ 0.000 & 0.338 $\pm$ 0.000 & 0.150 $\pm$ 0.000 & 0.086 $\pm$ 0.000 & 0.126 $\pm$ 0.000 & 0.680 $\pm$ 0.000 \\
\midrule
AxonAD (ours) & \textbf{0.285 $\pm$ 0.014} & \textbf{0.702 $\pm$ 0.011} & \textbf{0.157 $\pm$ 0.012} & \textbf{0.634 $\pm$ 0.017} & 0.533 $\pm$ 0.016 & \textbf{0.420 $\pm$ 0.019} & \textbf{0.328 $\pm$ 0.014} & 0.715 $\pm$ 0.024 \\
\bottomrule
\end{tabular}%
}
\end{table}

Table~\ref{tab:telemetry} reports results on the proprietary telemetry stream. AxonAD achieves the strongest threshold-free metrics by a wide margin, with AUC-PR of 0.285 versus 0.128 for the next best method (SISVAE). The gains are especially pronounced on Event-F1 (0.420 vs 0.255) and Range-F1 (0.328 vs 0.262), indicating that AxonAD not only ranks anomalies more accurately but also localizes them better in time. The large gap is consistent with the prevalence of coordination breaks in this dataset: anomalies that alter cross-channel dependencies without producing large per-channel excursions are precisely the regime where query mismatch provides the most value.

\begin{table}[t]
\centering
\caption{TSB-AD multivariate benchmark (17 datasets, 180 series). Mean $\pm$ std over all series. Best per metric in \textbf{bold}.}
\label{tab:tsbad_multivariate}
\resizebox{\textwidth}{!}{%
\begin{tabular}{@{}lcccccccc@{}}
\toprule
\multirow{2}{*}{\textbf{Model}}
& \multicolumn{8}{c}{\textbf{TSB-AD (multivariate, 17 datasets, 180 time series)}} \\ \cmidrule(l){2-9}
& \textbf{AUC-PR} & \textbf{AUC-ROC} & \textbf{VUS-PR} & \textbf{VUS-ROC}
& \textbf{PA-F1} & \textbf{Event-F1} & \textbf{Range-F1} & \textbf{Affiliation-F1} \\
\midrule
VASP~\cite{von_schleinitz_vasp_2021} & 0.339 $\pm$ 0.319 & 0.762 $\pm$ 0.195 & 0.401 $\pm$ 0.338 & 0.809 $\pm$ 0.185 & 0.669 $\pm$ 0.318 & 0.520 $\pm$ 0.361 & 0.400 $\pm$ 0.260 & 0.849 $\pm$ 0.123 \\
OmniAnomaly~\cite{su_robust_2019} & 0.372 $\pm$ 0.341 & 0.744 $\pm$ 0.250 & 0.424 $\pm$ 0.354 & 0.777 $\pm$ 0.240 & 0.627 $\pm$ 0.354 & 0.528 $\pm$ 0.367 & 0.432 $\pm$ 0.292 & 0.841 $\pm$ 0.126 \\
WVAE~\cite{pereira_unsupervised_2019} & 0.354 $\pm$ 0.331 & 0.747 $\pm$ 0.248 & 0.413 $\pm$ 0.349 & 0.778 $\pm$ 0.248 & 0.576 $\pm$ 0.388 & 0.502 $\pm$ 0.383 & 0.365 $\pm$ 0.280 & 0.838 $\pm$ 0.137 \\
USAD~\cite{audibert_usad_2020} & 0.363 $\pm$ 0.339 & 0.738 $\pm$ 0.256 & 0.412 $\pm$ 0.350 & 0.771 $\pm$ 0.244 & 0.622 $\pm$ 0.355 & 0.519 $\pm$ 0.364 & 0.422 $\pm$ 0.288 & 0.837 $\pm$ 0.131 \\
SISVAE~\cite{li_anomaly_2021} & 0.323 $\pm$ 0.290 & 0.759 $\pm$ 0.234 & 0.372 $\pm$ 0.315 & 0.786 $\pm$ 0.227 & 0.551 $\pm$ 0.367 & 0.470 $\pm$ 0.355 & 0.369 $\pm$ 0.278 & 0.824 $\pm$ 0.129 \\
MAVAE~\cite{correia_ma-vae_2023} & 0.299 $\pm$ 0.297 & 0.697 $\pm$ 0.256 & 0.351 $\pm$ 0.322 & 0.728 $\pm$ 0.256 & 0.568 $\pm$ 0.372 & 0.463 $\pm$ 0.360 & 0.325 $\pm$ 0.264 & 0.812 $\pm$ 0.132 \\
VSVAE~\cite{pereira_unsupervised_2018} & 0.290 $\pm$ 0.286 & 0.709 $\pm$ 0.256 & 0.342 $\pm$ 0.321 & 0.734 $\pm$ 0.257 & 0.596 $\pm$ 0.355 & 0.487 $\pm$ 0.347 & 0.374 $\pm$ 0.254 & 0.841 $\pm$ 0.121 \\
M2N2~\cite{abrantes_competition_2025} & 0.319 $\pm$ 0.358 & 0.740 $\pm$ 0.198 & 0.323 $\pm$ 0.359 & 0.779 $\pm$ 0.183 & \textbf{0.876 $\pm$ 0.184} & 0.603 $\pm$ 0.372 & 0.282 $\pm$ 0.233 & 0.860 $\pm$ 0.118 \\
TranAD~\cite{tuli_tranad_2022} & 0.258 $\pm$ 0.318 & 0.675 $\pm$ 0.221 & 0.308 $\pm$ 0.347 & 0.742 $\pm$ 0.210 & 0.753 $\pm$ 0.314 & 0.530 $\pm$ 0.367 & 0.218 $\pm$ 0.154 & 0.826 $\pm$ 0.125 \\
TFTResidual~\cite{lim_temporal_2021} & 0.250 $\pm$ 0.313 & 0.710 $\pm$ 0.210 & 0.308 $\pm$ 0.338 & 0.777 $\pm$ 0.186 & 0.746 $\pm$ 0.318 & 0.472 $\pm$ 0.362 & 0.207 $\pm$ 0.161 & 0.846 $\pm$ 0.114 \\
TimesNet~\cite{wu_timesnet_2023} & 0.201 $\pm$ 0.246 & 0.618 $\pm$ 0.279 & 0.271 $\pm$ 0.297 & 0.686 $\pm$ 0.277 & 0.750 $\pm$ 0.292 & 0.427 $\pm$ 0.354 & 0.176 $\pm$ 0.129 & 0.821 $\pm$ 0.117 \\
IForest~\cite{liu_isolation_2008} & 0.210 $\pm$ 0.232 & 0.704 $\pm$ 0.191 & 0.253 $\pm$ 0.260 & 0.750 $\pm$ 0.184 & 0.655 $\pm$ 0.335 & 0.403 $\pm$ 0.322 & 0.243 $\pm$ 0.178 & 0.801 $\pm$ 0.110 \\
LSTMAD~\cite{malhotra_lstm_2015} & 0.248 $\pm$ 0.328 & 0.597 $\pm$ 0.337 & 0.245 $\pm$ 0.329 & 0.626 $\pm$ 0.343 & 0.657 $\pm$ 0.412 & 0.507 $\pm$ 0.413 & 0.198 $\pm$ 0.175 & 0.701 $\pm$ 0.350 \\
\midrule
AxonAD (ours) & \textbf{0.437 $\pm$ 0.323} & \textbf{0.825 $\pm$ 0.169} & \textbf{0.493 $\pm$ 0.325} & \textbf{0.859 $\pm$ 0.146} & 0.698 $\pm$ 0.316 & 0.600 $\pm$ 0.336 & \textbf{0.471 $\pm$ 0.290} & 0.860 $\pm$ 0.132 \\
\bottomrule
\end{tabular}%
}
\end{table}

Table~\ref{tab:tsbad_multivariate} shows that these gains generalize beyond telemetry. On the TSB-AD multivariate suite, AxonAD achieves the highest mean AUC-PR (0.437), VUS-PR (0.493), and Range-F1 (0.471). M2N2 leads on PA-F1, and VASP and OmniAnomaly are competitive on Affiliation-F1, but all three rank below AxonAD on threshold-free metrics. Classical detectors achieve moderate AUC-ROC but lower AUC-PR and range-aware scores. Transformer-based detectors are competitive on subsets of series but show lower mean ranking in aggregate.

\begin{figure}[t]
\centering
\begin{tikzpicture}
\begin{groupplot}[
  group style={group size=2 by 1, horizontal sep=1.0cm},
  width=0.45\linewidth,
  height=6.0cm,
  y dir=reverse,
  ytick=data,
  tick label style={font=\scriptsize},
  label style={font=\scriptsize},
  title style={font=\scriptsize},
  symbolic y coords={
    OmniAnomaly,USAD,VASP,M2N2,TranAD,TFTResidual,TimesNet
  },
  xmajorgrids=true,
  grid style={dotted, black!15},
  legend style={
      at={(0.5,-0.18)},
      anchor=north,
      legend columns=2,
      font=\scriptsize,
  }
]

\nextgroupplot[
  title={AUC-PR: wins/losses (out of 180)},
  xbar stacked,
  xmin=0, xmax=180,
  xlabel={\# series},
]
\addplot+[xbar, fill=pastgray!30, draw=pastgray!60] coordinates {
  (59,OmniAnomaly) (53,USAD) (52,VASP) (43,M2N2) (35,TranAD) (24,TFTResidual) (21,TimesNet)
};
\addlegendentry{baseline wins}
\addplot+[xbar, fill=axonblue!55, draw=axonblue!80] coordinates {
  (121,OmniAnomaly) (127,USAD) (128,VASP) (137,M2N2) (145,TranAD) (156,TFTResidual) (159,TimesNet)
};
\addlegendentry{AxonAD wins}

\nextgroupplot[
  title={Median $\Delta$ (baseline $-$ AxonAD)},
  xmin=-0.25, xmax=0.04,
  xlabel={AUC-PR difference},
  yticklabels=\empty,
]
\draw[thick, dashed, black!30] (axis cs:0,OmniAnomaly) -- (axis cs:0,TimesNet);
\addplot+[
  only marks,
  mark=*,
  mark size=2.2pt,
  axonblue,
] coordinates {
  (-0.0190,OmniAnomaly)
  (-0.0275,USAD)
  (-0.0449,VASP)
  (-0.0955,M2N2)
  (-0.1080,TranAD)
  (-0.1252,TFTResidual)
  (-0.1916,TimesNet)
};
\draw[thick, axonblue!50] (axis cs:0,OmniAnomaly) -- (axis cs:-0.0190,OmniAnomaly);
\draw[thick, axonblue!50] (axis cs:0,USAD) -- (axis cs:-0.0275,USAD);
\draw[thick, axonblue!50] (axis cs:0,VASP) -- (axis cs:-0.0449,VASP);
\draw[thick, axonblue!50] (axis cs:0,M2N2) -- (axis cs:-0.0955,M2N2);
\draw[thick, axonblue!50] (axis cs:0,TranAD) -- (axis cs:-0.1080,TranAD);
\draw[thick, axonblue!50] (axis cs:0,TFTResidual) -- (axis cs:-0.1252,TFTResidual);
\draw[thick, axonblue!50] (axis cs:0,TimesNet) -- (axis cs:-0.1916,TimesNet);
\node[anchor=south west, font=\scriptsize, axonblue!80!black] at (axis cs:0.003,TimesNet) {$\leftarrow$ favors AxonAD};

\end{groupplot}
\end{tikzpicture}
\caption{Paired AUC-PR comparison on TSB-AD multivariate ($n=180$). Left: win/loss counts. Right: median paired difference with lollipop connectors from zero. All paired Wilcoxon tests yield $p < 10^{-4}$ with entirely negative 95\% bootstrap CIs (full statistics in the Appendix).}
\label{fig:paired_ranking_aucpr}
\end{figure}
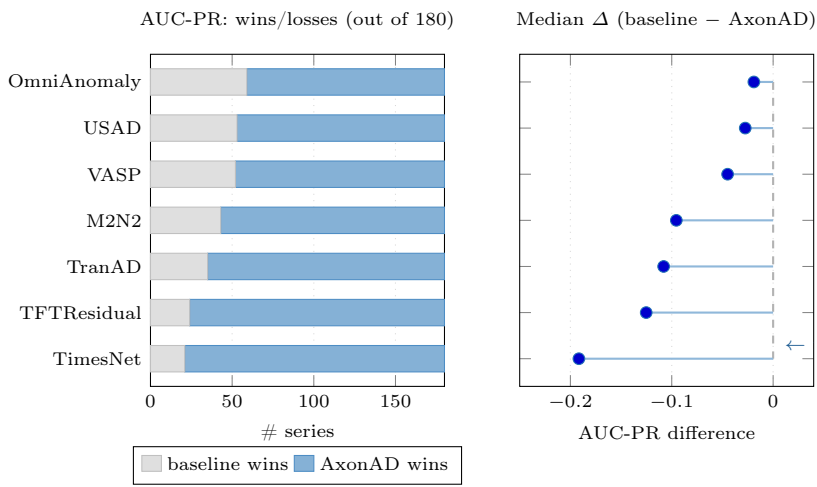

Figure~\ref{fig:paired_ranking_aucpr} confirms that improvements are broadly distributed: AxonAD wins on a clear majority of the 180 series against every baseline, with all paired Wilcoxon signed-rank tests yielding $p < 10^{-4}$.

\section{Ablation Studies}
\label{sec:ablation}


\begin{table}[t]
\centering
\caption{AxonAD ablation on the TSB-AD multivariate tuning subset (20 series). Mean $\pm$ std. Best per metric in \textbf{bold}. Rows are grouped to match the discussion order below.}
\label{tab:ablation}
\resizebox{\textwidth}{!}{%
\begin{tabular}{@{}lcccccccc@{}}
\toprule
\multirow{2}{*}{\textbf{Variant}}
& \multicolumn{8}{c}{\textbf{TSB-AD (multivariate, ablation subset)}} \\ \cmidrule(l){2-9}
& \textbf{AUC-PR} & \textbf{AUC-ROC} & \textbf{VUS-PR} & \textbf{VUS-ROC}
& \textbf{PA-F1} & \textbf{Event-F1} & \textbf{Range-F1} & \textbf{Affiliation-F1} \\
\midrule
Base                  & \textbf{0.558 $\pm$ 0.285} & \textbf{0.861 $\pm$ 0.137} & \textbf{0.658 $\pm$ 0.301} & 0.915 $\pm$ 0.102 & \textbf{0.855 $\pm$ 0.248} & \textbf{0.773 $\pm$ 0.262} & \textbf{0.564 $\pm$ 0.263} & \textbf{0.904 $\pm$ 0.123} \\
Recon only            & 0.511 $\pm$ 0.330 & 0.820 $\pm$ 0.218 & 0.603 $\pm$ 0.366 & 0.858 $\pm$ 0.223 & 0.728 $\pm$ 0.325 & 0.656 $\pm$ 0.322 & 0.541 $\pm$ 0.303 & 0.856 $\pm$ 0.142 \\
Score MSE             & 0.513 $\pm$ 0.327 & 0.828 $\pm$ 0.204 & 0.604 $\pm$ 0.365 & 0.868 $\pm$ 0.208 & 0.730 $\pm$ 0.317 & 0.652 $\pm$ 0.311 & 0.534 $\pm$ 0.293 & 0.852 $\pm$ 0.139 \\
JEPA only, Q          & 0.413 $\pm$ 0.317 & 0.764 $\pm$ 0.200 & 0.533 $\pm$ 0.353 & 0.846 $\pm$ 0.177 & 0.822 $\pm$ 0.283 & 0.683 $\pm$ 0.321 & 0.396 $\pm$ 0.248 & 0.892 $\pm$ 0.114 \\
\midrule
Score MSE+JEPA KL     & 0.554 $\pm$ 0.285 & 0.860 $\pm$ 0.137 & 0.655 $\pm$ 0.300 & \textbf{0.916 $\pm$ 0.100} & 0.854 $\pm$ 0.248 & 0.772 $\pm$ 0.262 & 0.560 $\pm$ 0.266 & 0.896 $\pm$ 0.127 \\
\midrule
EMA 0                 & 0.534 $\pm$ 0.302 & 0.855 $\pm$ 0.143 & 0.636 $\pm$ 0.319 & 0.908 $\pm$ 0.113 & 0.818 $\pm$ 0.250 & 0.722 $\pm$ 0.266 & 0.560 $\pm$ 0.258 & 0.873 $\pm$ 0.134 \\
EMA 0.99              & 0.510 $\pm$ 0.325 & 0.859 $\pm$ 0.137 & 0.608 $\pm$ 0.350 & 0.910 $\pm$ 0.105 & 0.824 $\pm$ 0.246 & 0.700 $\pm$ 0.269 & \textbf{0.564 $\pm$ 0.263} & 0.883 $\pm$ 0.127 \\
EMA 0.999             & 0.527 $\pm$ 0.310 & 0.856 $\pm$ 0.146 & 0.636 $\pm$ 0.322 & 0.913 $\pm$ 0.114 & 0.804 $\pm$ 0.249 & 0.724 $\pm$ 0.269 & 0.556 $\pm$ 0.261 & 0.882 $\pm$ 0.134 \\
Mask 0.8              & 0.533 $\pm$ 0.306 & 0.859 $\pm$ 0.142 & 0.631 $\pm$ 0.329 & 0.911 $\pm$ 0.111 & 0.808 $\pm$ 0.251 & 0.703 $\pm$ 0.271 & 0.547 $\pm$ 0.263 & 0.864 $\pm$ 0.137 \\
\midrule
Heads=4               & 0.525 $\pm$ 0.306 & 0.856 $\pm$ 0.139 & 0.630 $\pm$ 0.323 & 0.914 $\pm$ 0.101 & 0.808 $\pm$ 0.247 & 0.711 $\pm$ 0.258 & 0.531 $\pm$ 0.272 & 0.876 $\pm$ 0.128 \\
$D{=}64$                  & 0.516 $\pm$ 0.334 & 0.836 $\pm$ 0.183 & 0.613 $\pm$ 0.357 & 0.896 $\pm$ 0.147 & 0.796 $\pm$ 0.260 & 0.692 $\pm$ 0.289 & 0.553 $\pm$ 0.267 & 0.860 $\pm$ 0.142 \\
Horizon 25            & 0.502 $\pm$ 0.328 & 0.854 $\pm$ 0.150 & 0.599 $\pm$ 0.361 & 0.907 $\pm$ 0.118 & 0.821 $\pm$ 0.250 & 0.676 $\pm$ 0.294 & 0.517 $\pm$ 0.304 & 0.881 $\pm$ 0.128 \\
\midrule
Predict keys          & 0.405 $\pm$ 0.332 & 0.735 $\pm$ 0.233 & 0.495 $\pm$ 0.369 & 0.803 $\pm$ 0.227 & 0.748 $\pm$ 0.312 & 0.622 $\pm$ 0.359 & 0.462 $\pm$ 0.279 & 0.857 $\pm$ 0.126 \\
Predict values        & 0.405 $\pm$ 0.332 & 0.736 $\pm$ 0.232 & 0.496 $\pm$ 0.372 & 0.801 $\pm$ 0.228 & 0.744 $\pm$ 0.326 & 0.623 $\pm$ 0.358 & 0.439 $\pm$ 0.294 & 0.857 $\pm$ 0.126 \\
Predict attn map, Q   & 0.403 $\pm$ 0.334 & 0.754 $\pm$ 0.214 & 0.500 $\pm$ 0.378 & 0.832 $\pm$ 0.198 & 0.766 $\pm$ 0.307 & 0.630 $\pm$ 0.361 & 0.379 $\pm$ 0.251 & 0.870 $\pm$ 0.133 \\
Predict attn map, QK  & 0.388 $\pm$ 0.369 & 0.757 $\pm$ 0.228 & 0.486 $\pm$ 0.395 & 0.826 $\pm$ 0.210 & 0.709 $\pm$ 0.334 & 0.553 $\pm$ 0.365 & 0.402 $\pm$ 0.286 & 0.844 $\pm$ 0.132 \\
Predict hidden state  & 0.400 $\pm$ 0.337 & 0.742 $\pm$ 0.234 & 0.481 $\pm$ 0.371 & 0.815 $\pm$ 0.221 & 0.720 $\pm$ 0.337 & 0.604 $\pm$ 0.360 & 0.428 $\pm$ 0.260 & 0.856 $\pm$ 0.116 \\
\bottomrule
\end{tabular}%
}
\end{table}

Table~\ref{tab:ablation} reports ablations on the TSB-AD multivariate tuning subset (20 series)
under the official protocol.  All variants share identical preprocessing, windowing, and metric
computation.  Rows are grouped by the design dimension under study and discussed in that order below.

\myParagraph{Scoring components}
The base configuration (\emph{Base}) achieves the stron\-gest balanced profile across ranking and
localization metrics.  Removing the query branch at inference and using $S = \mathrm{rz}(d_{\mathrm{rec}})$
alone (\emph{Recon only}) reduces VUS-PR by 0.055 and Event-F1 by 0.117.  Retaining both branches
but replacing cosine mismatch with MSE distance in query space (\emph{Score MSE}) yields a similar
drop, indicating that the cosine formulation matters beyond simply combining two scores.  Using the
query signal alone (\emph{JEPA only, Q}) reduces AUC-PR by 0.145 and AUC-ROC by 0.097 despite
retaining competitive PA-F1, confirming that reconstruction is necessary for reliable ranking across
all anomaly types.  The cosine-based combined score therefore yields the most reliable behavior
across metric families.

\myParagraph{KL tail}
Adding attention divergence on top of the default score (\emph{Score MSE+JEPA KL}) yields no
consistent improvement over \emph{Base} on any metric.  We treat the KL tail as a diagnostic signal
only and exclude it from the default scoring pipeline.

\myParagraph{EMA and masking}
Removing the EMA target encoder entirely (\emph{EMA 0}, i.e.\ $m=0$) reduces AUC-PR by 0.024 and
Event-F1 by 0.051.  Moderate momentum (\emph{EMA 0.99}, $m=0.99$) incurs a similar AUC-PR penalty
of 0.048, while very high momentum (\emph{EMA 0.999}, $m=0.999$) likewise degrades ranking; both
extremes confirm that the default $m=0.9$ strikes the right balance between target stability and
responsiveness to online updates.  Increasing the masking ratio to 0.8 (\emph{Mask 0.8}) similarly
reduces AUC-PR and Event-F1, indicating that overly aggressive masking makes the predictive task too
hard during training.

\myParagraph{Capacity and horizon}
Reducing the number of attention heads from 8 to 4 (\emph{Heads=4}) lowers AUC-PR by 0.033 with a
smaller effect on localization metrics.  Halving the model dimension from 128 to 64 (\emph{$D{=}64$})
reduces AUC-PR by 0.042 and AUC-ROC by 0.025.  Increasing the forecast horizon to $s=25$
(\emph{Horizon 25}) reduces AUC-PR by 0.056, consistent with a harder prediction task introducing
more score variance at inference.

\myParagraph{Prediction target}
Predicting keys (\emph{Predict keys}), values (\emph{Predict values}), attention maps scored with
query inputs only (\emph{Predict attn map, Q}), attention maps scored with both query and key inputs
(\emph{Predict attn map, QK}), or intermediate hidden states (\emph{Predict hidden state}) is
consistently inferior to predicting query vectors across all ranking and localization metrics,
supporting the design choice of query prediction as the supervision and scoring target.


\begin{figure}[t]
\centering
\begin{minipage}{0.48\linewidth}
\centering
\begin{tikzpicture}
\begin{axis}[
  width=\linewidth,
  height=4.8cm,
  ymin=0.35, ymax=0.75,
  xmin=0, xmax=26,
  title={Forecast horizon $s$},
  xlabel={$s$},
  ylabel={metric value},
  xtick={1,3,5,25},
  xticklabels={1,3,5,25},
  xmajorgrids=true,
  ymajorgrids=true,
  grid style={dotted, black!15},
  tick label style={font=\scriptsize},
  label style={font=\scriptsize},
  title style={font=\scriptsize},
  legend style={font=\scriptsize, at={(0.98,0.02)}, anchor=south east, fill=white, fill opacity=0.9, draw=none},
]
\addplot+[thick, mark=*, mark size=2.0pt, axonblue] coordinates {(1,0.545) (3,0.532) (5,0.500) (25,0.517)};
\addlegendentry{AUC-PR}
\addplot+[thick, mark=square*, mark size=2.0pt, accentorange] coordinates {(1,0.648) (3,0.641) (5,0.613) (25,0.617)};
\addlegendentry{VUS-PR}
\addplot+[thick, mark=triangle*, mark size=2.0pt, anomred] coordinates {(1,0.553) (3,0.527) (5,0.520) (25,0.511)};
\addlegendentry{Range-F1}
\end{axis}
\end{tikzpicture}
\end{minipage}
\hfill
\begin{minipage}{0.48\linewidth}
\centering
\begin{tikzpicture}
\begin{axis}[
  width=\linewidth,
  height=4.8cm,
  ymin=0.35, ymax=0.75,
  xmin=0, xmax=21,
  title={Tail length $k$},
  xlabel={$k$},
  ylabel={metric value},
  xtick={3,5,10,20},
  xticklabels={3,5,10,20},
  xmajorgrids=true,
  ymajorgrids=true,
  grid style={dotted, black!15},
  tick label style={font=\scriptsize},
  label style={font=\scriptsize},
  title style={font=\scriptsize},
  legend style={font=\scriptsize, at={(0.98,0.02)}, anchor=south east, fill=white, fill opacity=0.9, draw=none},
]
\addplot+[thick, mark=*, mark size=2.0pt, axonblue] coordinates {(3,0.537) (5,0.534) (10,0.535) (20,0.524)};
\addlegendentry{AUC-PR}
\addplot+[thick, mark=square*, mark size=2.0pt, accentorange] coordinates {(3,0.641) (5,0.641) (10,0.631) (20,0.623)};
\addlegendentry{VUS-PR}
\addplot+[thick, mark=triangle*, mark size=2.0pt, anomred] coordinates {(3,0.509) (5,0.543) (10,0.564) (20,0.555)};
\addlegendentry{Range-F1}
\end{axis}
\end{tikzpicture}
\end{minipage}
\caption{Sensitivity to forecast horizon $s$ (left) and tail aggregation length $k$ (right) on the tuning subset.}
\label{fig:sensitivity}
\end{figure}
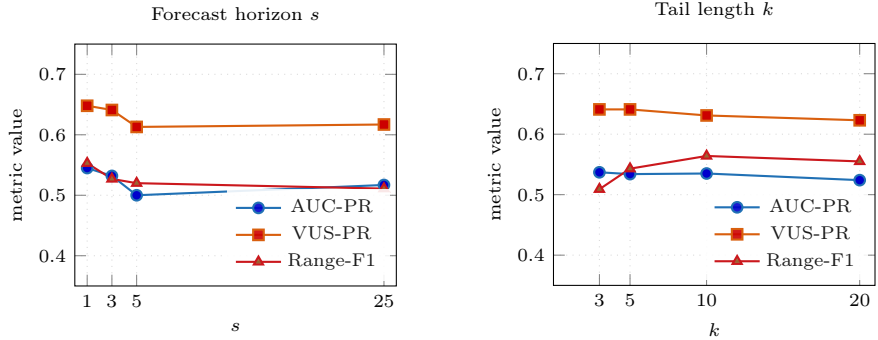

\myParagraph{Parameter sensitivity}
Figure~\ref{fig:sensitivity} shows sensitivity to the forecast horizon $s$ and the tail length $k$. Performance peaks at $s=1$ (AUC-PR 0.545, Range-F1 0.553) and is generally lower for larger horizons, as a harder prediction task increases score variance. For tail length, threshold-free ranking is stable across $k\in\{3,5,10,20\}$ (AUC-PR in $[0.524,0.537]$), while Range-F1 peaks at $k=10$, suggesting that $k$ primarily controls temporal smoothing.

\myParagraph{Mechanistic diagnostics}
To verify that query mismatch captures meaningful attention structure rather than noise, we run a descriptive analysis on the tuning subset (not used for model selection). Spearman correlation between query deviation magnitude $\|\Delta Q\|$ and tail KL divergence $\mathrm{KL}(A_{\mathrm{tgt}}\|A_{\mathrm{pred}})$, where $A_{\mathrm{tgt}}$ and $A_{\mathrm{pred}}$ denote attention weights from EMA target and predicted queries respectively, is frequently positive (median $\rho=0.677$, $\rho\ge 0.50$ in 15 of 20 series). This confirms that query mismatch tracks genuine attention redistribution. Tail attention entropy is nondegenerate (range 3.18 to 4.53), ruling out collapsed attention as a confound.

The window-level correlation between reconstruction error and query mismatch is small (median $\rho=0.211$). Among anomalous windows, the fraction with high query mismatch but low reconstruction error is 0.192, and the reverse is 0.095. Both regimes occur in most series, and combining components improves AUC-PR over the best single component in 8 of 20 series. This supports the coverage interpretation underlying the combined score.

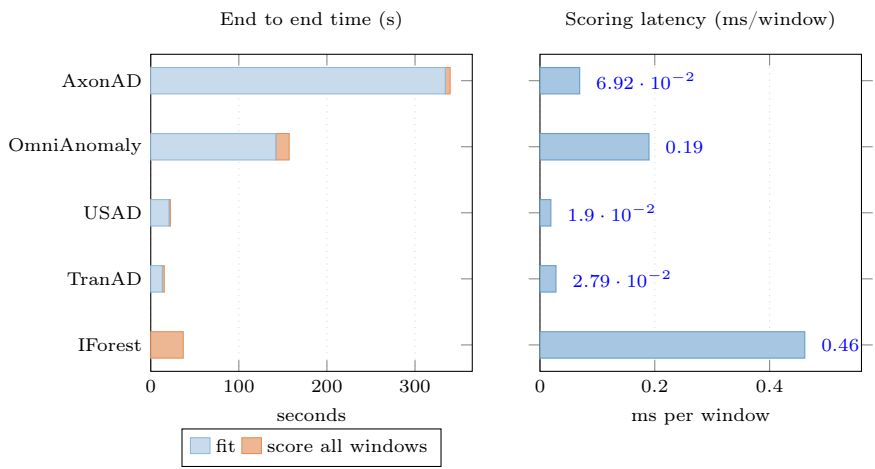
\begin{figure}[t]
\centering
\begin{tikzpicture}
\begin{groupplot}[
  group style={group size=2 by 1, horizontal sep=0.9cm},
  width=0.48\linewidth,
  height=5.8cm,
  y dir=reverse,
  ytick=data,
  tick label style={font=\scriptsize},
  label style={font=\scriptsize},
  title style={font=\scriptsize},
legend style={
  at={(0.5,-0.18)},
  anchor=north,
  legend columns=2,
  font=\scriptsize
},
  symbolic y coords={AxonAD,OmniAnomaly,USAD,TranAD,IForest},
  xmajorgrids=true,
  grid style={dotted, black!15},
]

\nextgroupplot[
  title={End to end time (s)},
  xbar stacked,
  xmin=0, xmax=365,
  xlabel={seconds},
]
\addplot+[xbar, fill=axonblue!25, draw=axonblue!50] coordinates {
  (334.34,AxonAD)
  (141.96,OmniAnomaly)
  (20.72,USAD)
  (13.11,TranAD)
  (0,IForest)
};
\addlegendentry{fit}
\addplot+[xbar, fill=accentorange!45, draw=accentorange!70] coordinates {
  (5.53,AxonAD)
  (15.17,OmniAnomaly)
  (1.52,USAD)
  (2.23,TranAD)
  (36.90,IForest)
};
\addlegendentry{score all windows}

\nextgroupplot[
  title={Scoring latency (ms/window)},
  xbar,
  xmin=0, xmax=0.56,
  xlabel={ms per window},
  yticklabels=\empty,
  nodes near coords,
  every node near coord/.append style={font=\scriptsize, xshift=3pt},
]
\addplot+[xbar, fill=axonblue!40, draw=axonblue!70] coordinates {
  (0.0692,AxonAD)
  (0.1899,OmniAnomaly)
  (0.0190,USAD)
  (0.0279,TranAD)
  (0.4613,IForest)
};

\end{groupplot}
\end{tikzpicture}
\caption{Runtime on the telemetry stream (80{,}000 samples, stride 1). Left: wall-clock time split into fitting and scoring. Right: per window scoring latency. Classical baselines have no iterative training, so their time is entirely attributed to scoring.}
\label{fig:runtime}
\end{figure}

\myParagraph{Runtime}
Figure~\ref{fig:runtime} profiles runtime on the telemetry stream. AxonAD has the highest fitting cost (334\,s) but achieves per window scoring latency of 0.069\,ms, lower than OmniAnomaly (0.190\,ms) and Isolation Forest (0.461\,ms). The fitting cost reflects iterative gradient-based training and is amortized at deployment. For a fleet monitoring pipeline processing windows at 10\,Hz, the 0.069\,ms latency leaves ample margin for real-time operation.

\section{Conclusion}

AxonAD detects multivariate time series anomalies by monitoring the predictability of attention query vectors. A bidirectional reconstruction pathway is coupled with a history-only predictor trained via masked EMA distillation in query space, producing a query mismatch signal that complements reconstruction residuals and responds to structural dependency shifts. Robust standardization of both components enables reliable score combination across heterogeneous datasets without label-based calibration.

On proprietary in-vehicle telemetry, where coordination breaks between steering, acceleration, and powertrain channels are the dominant fault mode, AxonAD improves AUC-PR by $2.2\times$ over the next best baseline and Event-F1 by $1.6\times$. These gains transfer to the multivariate TSB-AD benchmark (17 datasets, 180 series), where AxonAD leads on threshold-free ranking and range-aware localization. Ablations establish that query prediction outperforms alternative predictive targets and that combining both scores is necessary for the best aggregate performance. The low per window inference latency (0.069\,ms) and the absence of label-based threshold tuning support integration into streaming vehicle monitoring pipelines.

\bibliographystyle{splncs04}
\bibliography{references}

\renewcommand{\thesection}{S\arabic{section}}
\renewcommand{\thetable}{S\arabic{table}}
\renewcommand{\thefigure}{S\arabic{figure}}

\title{Predictable Query Dynamics for Time-Series Anomaly Detection}
\author{Supplementary Material} 
\institute{}              
\maketitle
\section{Additional Tables}
\begin{table}[htbp]
\centering
\caption{TSB-AD multivariate benchmark (17 datasets, 180 time series).
Mean $\pm$ standard deviation over all evaluated series.
Best result per metric in \textbf{bold}.}
\label{tab:tsbad_multivariate_appendix}
\resizebox{\textwidth}{!}{%
\begin{tabular}{@{}lcccccccc@{}}
\toprule
\multirow{2}{*}{\textbf{Model}}
& \multicolumn{8}{c}{\textbf{TSB-AD (multivariate, 17 datasets, 180 time series)}} \\ \cmidrule(l){2-9}
& \textbf{AUC-PR} & \textbf{AUC-ROC} & \textbf{VUS-PR} & \textbf{VUS-ROC}
& \textbf{PA-F1} & \textbf{Event-F1} & \textbf{Range-F1} & \textbf{Affiliation-F1} \\
\midrule
VASP & 0.339 $\pm$ 0.319 & 0.762 $\pm$ 0.195 & 0.401 $\pm$ 0.338 & 0.809 $\pm$ 0.185 & 0.669 $\pm$ 0.318 & 0.520 $\pm$ 0.361 & 0.400 $\pm$ 0.260 & 0.849 $\pm$ 0.123 \\
OmniAnomaly & 0.372 $\pm$ 0.341 & 0.744 $\pm$ 0.250 & 0.424 $\pm$ 0.354 & 0.777 $\pm$ 0.240 & 0.627 $\pm$ 0.354 & 0.528 $\pm$ 0.367 & 0.432 $\pm$ 0.292 & 0.841 $\pm$ 0.126 \\
WVAE & 0.354 $\pm$ 0.331 & 0.747 $\pm$ 0.248 & 0.413 $\pm$ 0.349 & 0.778 $\pm$ 0.248 & 0.576 $\pm$ 0.388 & 0.502 $\pm$ 0.383 & 0.365 $\pm$ 0.280 & 0.838 $\pm$ 0.137 \\
USAD & 0.363 $\pm$ 0.339 & 0.738 $\pm$ 0.256 & 0.412 $\pm$ 0.350 & 0.771 $\pm$ 0.244 & 0.622 $\pm$ 0.355 & 0.519 $\pm$ 0.364 & 0.422 $\pm$ 0.288 & 0.837 $\pm$ 0.131 \\
SISVAE & 0.323 $\pm$ 0.290 & 0.759 $\pm$ 0.234 & 0.372 $\pm$ 0.315 & 0.786 $\pm$ 0.227 & 0.551 $\pm$ 0.367 & 0.470 $\pm$ 0.355 & 0.369 $\pm$ 0.278 & 0.824 $\pm$ 0.129 \\
OFA & 0.300 $\pm$ 0.300 & 0.639 $\pm$ 0.289 & 0.367 $\pm$ 0.342 & 0.694 $\pm$ 0.286 & 0.675 $\pm$ 0.305 & 0.517 $\pm$ 0.372 & 0.360 $\pm$ 0.251 & 0.833 $\pm$ 0.126 \\
CNN & 0.347 $\pm$ 0.356 & 0.770 $\pm$ 0.176 & 0.352 $\pm$ 0.359 & 0.807 $\pm$ 0.164 & 0.828 $\pm$ 0.266 & \textbf{0.643 $\pm$ 0.366} & 0.301 $\pm$ 0.240 & \textbf{0.866 $\pm$ 0.120} \\
MAVAE & 0.299 $\pm$ 0.297 & 0.697 $\pm$ 0.256 & 0.351 $\pm$ 0.322 & 0.728 $\pm$ 0.256 & 0.568 $\pm$ 0.372 & 0.463 $\pm$ 0.360 & 0.325 $\pm$ 0.264 & 0.812 $\pm$ 0.132 \\
VSVAE & 0.290 $\pm$ 0.286 & 0.709 $\pm$ 0.256 & 0.342 $\pm$ 0.321 & 0.734 $\pm$ 0.257 & 0.596 $\pm$ 0.355 & 0.487 $\pm$ 0.347 & 0.374 $\pm$ 0.254 & 0.841 $\pm$ 0.121 \\
GDN & 0.272 $\pm$ 0.305 & 0.738 $\pm$ 0.193 & 0.332 $\pm$ 0.329 & 0.802 $\pm$ 0.175 & 0.756 $\pm$ 0.310 & 0.499 $\pm$ 0.364 & 0.208 $\pm$ 0.134 & 0.846 $\pm$ 0.119 \\
M2N2 & 0.319 $\pm$ 0.358 & 0.740 $\pm$ 0.198 & 0.323 $\pm$ 0.359 & 0.779 $\pm$ 0.183 & \textbf{0.876 $\pm$ 0.184} & 0.603 $\pm$ 0.372 & 0.282 $\pm$ 0.233 & 0.860 $\pm$ 0.118 \\
TranAD & 0.258 $\pm$ 0.318 & 0.675 $\pm$ 0.221 & 0.308 $\pm$ 0.347 & 0.742 $\pm$ 0.210 & 0.753 $\pm$ 0.314 & 0.530 $\pm$ 0.367 & 0.218 $\pm$ 0.154 & 0.826 $\pm$ 0.125 \\
TFTResidual & 0.250 $\pm$ 0.313 & 0.710 $\pm$ 0.210 & 0.308 $\pm$ 0.338 & 0.777 $\pm$ 0.186 & 0.746 $\pm$ 0.318 & 0.472 $\pm$ 0.362 & 0.207 $\pm$ 0.161 & 0.846 $\pm$ 0.114 \\
KMeansAD & 0.252 $\pm$ 0.267 & 0.691 $\pm$ 0.202 & 0.296 $\pm$ 0.302 & 0.732 $\pm$ 0.195 & 0.675 $\pm$ 0.309 & 0.483 $\pm$ 0.364 & 0.326 $\pm$ 0.235 & 0.819 $\pm$ 0.126 \\
AutoEncoder & 0.294 $\pm$ 0.373 & 0.669 $\pm$ 0.212 & 0.295 $\pm$ 0.368 & 0.691 $\pm$ 0.207 & 0.597 $\pm$ 0.357 & 0.439 $\pm$ 0.411 & 0.283 $\pm$ 0.297 & 0.800 $\pm$ 0.139 \\
PCA & 0.242 $\pm$ 0.293 & 0.676 $\pm$ 0.238 & 0.277 $\pm$ 0.307 & 0.712 $\pm$ 0.223 & 0.514 $\pm$ 0.340 & 0.370 $\pm$ 0.329 & 0.325 $\pm$ 0.263 & 0.789 $\pm$ 0.122 \\
TimesNet & 0.201 $\pm$ 0.246 & 0.618 $\pm$ 0.279 & 0.271 $\pm$ 0.297 & 0.686 $\pm$ 0.277 & 0.750 $\pm$ 0.292 & 0.427 $\pm$ 0.354 & 0.176 $\pm$ 0.129 & 0.821 $\pm$ 0.117 \\
FITS & 0.197 $\pm$ 0.253 & 0.611 $\pm$ 0.271 & 0.267 $\pm$ 0.300 & 0.686 $\pm$ 0.274 & 0.763 $\pm$ 0.281 & 0.422 $\pm$ 0.342 & 0.181 $\pm$ 0.131 & 0.816 $\pm$ 0.115 \\
Donut & 0.213 $\pm$ 0.270 & 0.627 $\pm$ 0.239 & 0.262 $\pm$ 0.308 & 0.693 $\pm$ 0.237 & 0.525 $\pm$ 0.414 & 0.406 $\pm$ 0.385 & 0.180 $\pm$ 0.167 & 0.769 $\pm$ 0.200 \\
CBLOF & 0.263 $\pm$ 0.344 & 0.664 $\pm$ 0.206 & 0.260 $\pm$ 0.341 & 0.697 $\pm$ 0.193 & 0.648 $\pm$ 0.340 & 0.448 $\pm$ 0.412 & 0.302 $\pm$ 0.283 & 0.811 $\pm$ 0.149 \\
IForest & 0.210 $\pm$ 0.232 & 0.704 $\pm$ 0.191 & 0.253 $\pm$ 0.260 & 0.750 $\pm$ 0.184 & 0.655 $\pm$ 0.335 & 0.403 $\pm$ 0.322 & 0.243 $\pm$ 0.178 & 0.801 $\pm$ 0.110 \\
LSTMAD & 0.248 $\pm$ 0.328 & 0.597 $\pm$ 0.337 & 0.245 $\pm$ 0.329 & 0.626 $\pm$ 0.343 & 0.657 $\pm$ 0.412 & 0.507 $\pm$ 0.413 & 0.198 $\pm$ 0.175 & 0.701 $\pm$ 0.350 \\
RobustPCA & 0.238 $\pm$ 0.349 & 0.589 $\pm$ 0.241 & 0.238 $\pm$ 0.352 & 0.616 $\pm$ 0.235 & 0.573 $\pm$ 0.387 & 0.379 $\pm$ 0.385 & 0.332 $\pm$ 0.318 & 0.789 $\pm$ 0.135 \\
EIF & 0.186 $\pm$ 0.218 & 0.667 $\pm$ 0.174 & 0.210 $\pm$ 0.258 & 0.708 $\pm$ 0.168 & 0.741 $\pm$ 0.291 & 0.438 $\pm$ 0.374 & 0.258 $\pm$ 0.230 & 0.812 $\pm$ 0.121 \\
COPOD & 0.205 $\pm$ 0.292 & 0.652 $\pm$ 0.185 & 0.203 $\pm$ 0.290 & 0.686 $\pm$ 0.177 & 0.717 $\pm$ 0.320 & 0.414 $\pm$ 0.364 & 0.242 $\pm$ 0.237 & 0.799 $\pm$ 0.128 \\
HBOS & 0.161 $\pm$ 0.196 & 0.633 $\pm$ 0.183 & 0.190 $\pm$ 0.247 & 0.672 $\pm$ 0.184 & 0.667 $\pm$ 0.331 & 0.399 $\pm$ 0.345 & 0.244 $\pm$ 0.225 & 0.796 $\pm$ 0.117 \\
KNN & 0.133 $\pm$ 0.156 & 0.499 $\pm$ 0.218 & 0.176 $\pm$ 0.220 & 0.580 $\pm$ 0.219 & 0.681 $\pm$ 0.366 & 0.447 $\pm$ 0.391 & 0.205 $\pm$ 0.149 & 0.791 $\pm$ 0.140 \\
LOF & 0.096 $\pm$ 0.092 & 0.534 $\pm$ 0.098 & 0.138 $\pm$ 0.192 & 0.597 $\pm$ 0.147 & 0.563 $\pm$ 0.368 & 0.325 $\pm$ 0.328 & 0.149 $\pm$ 0.143 & 0.764 $\pm$ 0.133 \\
AnomalyTransformer & 0.068 $\pm$ 0.060 & 0.506 $\pm$ 0.053 & 0.115 $\pm$ 0.184 & 0.538 $\pm$ 0.098 & 0.658 $\pm$ 0.359 & 0.361 $\pm$ 0.363 & 0.138 $\pm$ 0.121 & 0.737 $\pm$ 0.195 \\
\midrule
AxonAD (ours) & \textbf{0.437 $\pm$ 0.323} & \textbf{0.825 $\pm$ 0.169} & \textbf{0.493 $\pm$ 0.325} & \textbf{0.859 $\pm$ 0.146} & 0.698 $\pm$ 0.316 & 0.600 $\pm$ 0.336 & \textbf{0.471 $\pm$ 0.290} & 0.860 $\pm$ 0.132 \\
\bottomrule
\end{tabular}%
}
\end{table}

\begin{table}[htbp]
\centering
\caption{TSB-AD multivariate benchmark protocol on proprietary dataset.
Mean $\pm$ standard deviation over 4 random seeds.
Best result per metric in \textbf{bold}.}
\label{tab:tsbad_multivariate_seeded}
\resizebox{\textwidth}{!}{%
\begin{tabular}{@{}lcccccccc@{}}
\toprule
\multirow{2}{*}{\textbf{Model}}
& \multicolumn{8}{c}{\textbf{TSB-AD (Proprietary Dataset)}} \\ \cmidrule(l){2-9}
& \textbf{AUC-PR} & \textbf{AUC-ROC} & \textbf{VUS-PR} & \textbf{VUS-ROC}
& \textbf{PA-F1} & \textbf{Event-F1} & \textbf{Range-F1} & \textbf{Affiliation-F1} \\
\midrule
LSTMAD & 0.082 $\pm$ 0.004 & 0.651 $\pm$ 0.009 & 0.083 $\pm$ 0.004 & 0.624 $\pm$ 0.009 & 0.533 $\pm$ 0.014 & 0.255 $\pm$ 0.015 & 0.139 $\pm$ 0.006 & 0.723 $\pm$ 0.003 \\
SISVAE & 0.128 $\pm$ 0.030 & 0.586 $\pm$ 0.026 & 0.070 $\pm$ 0.012 & 0.504 $\pm$ 0.052 & 0.270 $\pm$ 0.100 & 0.231 $\pm$ 0.060 & 0.225 $\pm$ 0.054 & 0.699 $\pm$ 0.018 \\
TFTResidual & 0.071 $\pm$ 0.006 & 0.644 $\pm$ 0.025 & 0.070 $\pm$ 0.005 & 0.582 $\pm$ 0.018 & 0.424 $\pm$ 0.022 & 0.164 $\pm$ 0.019 & 0.110 $\pm$ 0.009 & \textbf{0.752 $\pm$ 0.026} \\
RobustPCA & 0.070 $\pm$ 0.000 & 0.634 $\pm$ 0.000 & 0.066 $\pm$ 0.000 & 0.570 $\pm$ 0.000 & 0.312 $\pm$ 0.000 & 0.162 $\pm$ 0.000 & 0.100 $\pm$ 0.000 & 0.680 $\pm$ 0.000 \\
VSVAE & 0.100 $\pm$ 0.005 & 0.617 $\pm$ 0.031 & 0.065 $\pm$ 0.005 & 0.535 $\pm$ 0.027 & 0.214 $\pm$ 0.048 & 0.188 $\pm$ 0.004 & 0.262 $\pm$ 0.037 & 0.730 $\pm$ 0.012 \\
M2N2 & 0.065 $\pm$ 0.001 & 0.596 $\pm$ 0.001 & 0.064 $\pm$ 0.001 & 0.553 $\pm$ 0.004 & 0.392 $\pm$ 0.022 & 0.196 $\pm$ 0.020 & 0.120 $\pm$ 0.009 & 0.680 $\pm$ 0.000 \\
OFA & 0.061 $\pm$ 0.002 & 0.597 $\pm$ 0.003 & 0.060 $\pm$ 0.001 & 0.542 $\pm$ 0.001 & 0.224 $\pm$ 0.010 & 0.109 $\pm$ 0.007 & 0.122 $\pm$ 0.015 & 0.680 $\pm$ 0.000 \\
MAVAE & 0.094 $\pm$ 0.006 & 0.561 $\pm$ 0.034 & 0.059 $\pm$ 0.006 & 0.487 $\pm$ 0.051 & 0.220 $\pm$ 0.076 & 0.199 $\pm$ 0.015 & 0.202 $\pm$ 0.017 & 0.680 $\pm$ 0.000 \\
KNN & 0.085 $\pm$ 0.000 & 0.563 $\pm$ 0.000 & 0.059 $\pm$ 0.000 & 0.506 $\pm$ 0.000 & 0.336 $\pm$ 0.000 & 0.231 $\pm$ 0.000 & 0.080 $\pm$ 0.000 & 0.680 $\pm$ 0.000 \\
CNN & 0.058 $\pm$ 0.002 & 0.568 $\pm$ 0.006 & 0.059 $\pm$ 0.002 & 0.524 $\pm$ 0.007 & 0.382 $\pm$ 0.033 & 0.186 $\pm$ 0.018 & 0.105 $\pm$ 0.004 & 0.681 $\pm$ 0.003 \\
WVAE & 0.087 $\pm$ 0.013 & 0.541 $\pm$ 0.043 & 0.057 $\pm$ 0.007 & 0.467 $\pm$ 0.050 & 0.249 $\pm$ 0.103 & 0.226 $\pm$ 0.038 & 0.163 $\pm$ 0.011 & 0.680 $\pm$ 0.000 \\
LOF & 0.055 $\pm$ 0.000 & 0.543 $\pm$ 0.000 & 0.056 $\pm$ 0.000 & 0.552 $\pm$ 0.000 & \textbf{0.561 $\pm$ 0.000} & 0.166 $\pm$ 0.000 & 0.124 $\pm$ 0.000 & 0.680 $\pm$ 0.000 \\
TimesNet & 0.055 $\pm$ 0.001 & 0.579 $\pm$ 0.003 & 0.056 $\pm$ 0.000 & 0.531 $\pm$ 0.003 & 0.306 $\pm$ 0.020 & 0.102 $\pm$ 0.004 & 0.092 $\pm$ 0.002 & 0.680 $\pm$ 0.000 \\
FITS & 0.050 $\pm$ 0.000 & 0.563 $\pm$ 0.000 & 0.053 $\pm$ 0.000 & 0.548 $\pm$ 0.001 & 0.451 $\pm$ 0.002 & 0.091 $\pm$ 0.001 & 0.071 $\pm$ 0.001 & 0.680 $\pm$ 0.000 \\
GDN & 0.052 $\pm$ 0.006 & 0.547 $\pm$ 0.027 & 0.052 $\pm$ 0.004 & 0.478 $\pm$ 0.027 & 0.349 $\pm$ 0.074 & 0.115 $\pm$ 0.028 & 0.087 $\pm$ 0.005 & 0.681 $\pm$ 0.004 \\
VASP & 0.050 $\pm$ 0.001 & 0.540 $\pm$ 0.014 & 0.051 $\pm$ 0.002 & 0.449 $\pm$ 0.016 & 0.190 $\pm$ 0.008 & 0.099 $\pm$ 0.004 & 0.119 $\pm$ 0.013 & 0.686 $\pm$ 0.008 \\
AutoEncoder & 0.047 $\pm$ 0.003 & 0.541 $\pm$ 0.018 & 0.051 $\pm$ 0.003 & 0.495 $\pm$ 0.022 & 0.185 $\pm$ 0.026 & 0.107 $\pm$ 0.005 & 0.084 $\pm$ 0.016 & 0.680 $\pm$ 0.000 \\
EIF & 0.049 $\pm$ 0.002 & 0.500 $\pm$ 0.006 & 0.047 $\pm$ 0.000 & 0.357 $\pm$ 0.010 & 0.224 $\pm$ 0.019 & 0.118 $\pm$ 0.014 & 0.192 $\pm$ 0.016 & 0.690 $\pm$ 0.007 \\
AnomalyTransformer & 0.045 $\pm$ 0.003 & 0.491 $\pm$ 0.019 & 0.047 $\pm$ 0.002 & 0.454 $\pm$ 0.032 & 0.379 $\pm$ 0.031 & 0.118 $\pm$ 0.033 & 0.064 $\pm$ 0.006 & 0.591 $\pm$ 0.086 \\
HBOS & 0.064 $\pm$ 0.000 & 0.479 $\pm$ 0.000 & 0.046 $\pm$ 0.000 & 0.349 $\pm$ 0.000 & 0.282 $\pm$ 0.000 & 0.171 $\pm$ 0.000 & 0.193 $\pm$ 0.000 & 0.701 $\pm$ 0.000 \\
KMeansAD & 0.042 $\pm$ 0.000 & 0.495 $\pm$ 0.000 & 0.046 $\pm$ 0.000 & 0.343 $\pm$ 0.000 & 0.161 $\pm$ 0.009 & 0.087 $\pm$ 0.002 & 0.098 $\pm$ 0.007 & 0.680 $\pm$ 0.000 \\
CBLOF & 0.041 $\pm$ 0.000 & 0.481 $\pm$ 0.000 & 0.044 $\pm$ 0.000 & 0.352 $\pm$ 0.000 & 0.155 $\pm$ 0.000 & 0.086 $\pm$ 0.000 & 0.073 $\pm$ 0.000 & 0.680 $\pm$ 0.000 \\
IForest & 0.041 $\pm$ 0.000 & 0.472 $\pm$ 0.000 & 0.044 $\pm$ 0.000 & 0.328 $\pm$ 0.000 & 0.140 $\pm$ 0.000 & 0.086 $\pm$ 0.000 & 0.195 $\pm$ 0.000 & 0.682 $\pm$ 0.000 \\
TranAD & 0.041 $\pm$ 0.000 & 0.470 $\pm$ 0.003 & 0.044 $\pm$ 0.000 & 0.417 $\pm$ 0.004 & 0.237 $\pm$ 0.003 & 0.086 $\pm$ 0.000 & 0.107 $\pm$ 0.007 & 0.680 $\pm$ 0.000 \\
USAD & 0.040 $\pm$ 0.001 & 0.470 $\pm$ 0.010 & 0.044 $\pm$ 0.001 & 0.371 $\pm$ 0.011 & 0.122 $\pm$ 0.005 & 0.087 $\pm$ 0.001 & 0.152 $\pm$ 0.033 & 0.682 $\pm$ 0.001 \\
PCA & 0.037 $\pm$ 0.000 & 0.447 $\pm$ 0.000 & 0.043 $\pm$ 0.000 & 0.377 $\pm$ 0.000 & 0.107 $\pm$ 0.000 & 0.092 $\pm$ 0.000 & 0.148 $\pm$ 0.000 & 0.684 $\pm$ 0.000 \\
OmniAnomaly & 0.041 $\pm$ 0.000 & 0.459 $\pm$ 0.000 & 0.043 $\pm$ 0.000 & 0.338 $\pm$ 0.000 & 0.150 $\pm$ 0.000 & 0.086 $\pm$ 0.000 & 0.126 $\pm$ 0.000 & 0.680 $\pm$ 0.000 \\
COPOD & 0.035 $\pm$ 0.000 & 0.433 $\pm$ 0.000 & 0.041 $\pm$ 0.000 & 0.368 $\pm$ 0.000 & 0.131 $\pm$ 0.000 & 0.090 $\pm$ 0.000 & 0.170 $\pm$ 0.000 & 0.710 $\pm$ 0.000 \\
Donut & 0.036 $\pm$ 0.001 & 0.443 $\pm$ 0.020 & 0.041 $\pm$ 0.001 & 0.386 $\pm$ 0.027 & 0.091 $\pm$ 0.006 & 0.086 $\pm$ 0.000 & 0.098 $\pm$ 0.059 & 0.680 $\pm$ 0.000
\\
\midrule
AxonAD (ours) & \textbf{0.285 $\pm$ 0.014} & \textbf{0.702 $\pm$ 0.011} & \textbf{0.157 $\pm$ 0.012} & \textbf{0.634 $\pm$ 0.017} & 0.533 $\pm$ 0.016 & \textbf{0.420 $\pm$ 0.019} & \textbf{0.328 $\pm$ 0.014} & 0.715 $\pm$ 0.024 \\
\bottomrule
\end{tabular}%
}
\end{table}


\begin{table}[htbp]
\centering
\caption{Pairwise comparison against \textbf{AxonAD} on TSB-AD multivariate (17 datasets, 180 time series).
For each baseline and metric we report: win-rate (wins/180), mean and median performance delta (\,$\Delta$\,), Wilcoxon $p$-value, and 95\% CI for $\Delta$.
Negative $\Delta$ indicates the baseline is worse than AxonAD (per your export).
Entries with non-significant Wilcoxon test ($p\ge 0.05$) are \textbf{bold}.}
\label{tab:tsbad_pairwise_vs_axonad}
\resizebox{\textwidth}{!}{%
\begin{tabular}{@{}lcccccccccc@{}}
\toprule
\multirow{2}{*}{\textbf{Model}}
& \multicolumn{5}{c}{\textbf{AUC-PR (vs AxonAD)}} 
& \multicolumn{5}{c}{\textbf{AUC-ROC (vs AxonAD)}} \\
\cmidrule(l){2-6}\cmidrule(l){7-11}
& \textbf{WR} & $\boldsymbol{\Delta}$ mean & $\boldsymbol{\Delta}$ med & $p$ & CI$_{95}$ 
& \textbf{WR} & $\boldsymbol{\Delta}$ mean & $\boldsymbol{\Delta}$ med & $p$ & CI$_{95}$ \\
\midrule
AnomalyTransformer & 0.039 & -0.3693 & -0.3203 & $1.24\!\times\!10^{-30}$ & [-0.4124,\,-0.3255] & 0.033 & -0.3185 & -0.3520 & $1.14\!\times\!10^{-29}$ & [-0.3432,\,-0.2919] \\
AutoEncoder        & 0.194 & -0.1427 & -0.1462 & $1.93\!\times\!10^{-11}$ & [-0.1969,\,-0.0882] & 0.172 & -0.1554 & -0.1776 & $2.33\!\times\!10^{-14}$ & [-0.1916,\,-0.1224] \\
CBLOF              & 0.178 & -0.1738 & -0.1506 & $2.19\!\times\!10^{-13}$ & [-0.2287,\,-0.1185] & 0.139 & -0.1610 & -0.1601 & $6.06\!\times\!10^{-16}$ & [-0.1949,\,-0.1268] \\
CNN                & 0.267 & -0.0899 & -0.0859 & $1.36\!\times\!10^{-8}$  & [-0.1388,\,-0.0393] & 0.228 & -0.0544 & -0.0555 & $2.24\!\times\!10^{-8}$  & [-0.0829,\,-0.0268] \\
COPOD              & 0.194 & -0.2324 & -0.2315 & $2.30\!\times\!10^{-14}$ & [-0.2903,\,-0.1716] & 0.133 & -0.1730 & -0.2094 & $2.19\!\times\!10^{-16}$ & [-0.2064,\,-0.1398] \\
Donut              & 0.117 & -0.2237 & -0.1705 & $2.11\!\times\!10^{-22}$ & [-0.2635,\,-0.1844] & 0.122 & -0.1978 & -0.1527 & $6.15\!\times\!10^{-23}$ & [-0.2329,\,-0.1657] \\
EIF                & 0.189 & -0.2505 & -0.2159 & $1.98\!\times\!10^{-17}$ & [-0.2998,\,-0.2029] & 0.150 & -0.1582 & -0.1657 & $1.57\!\times\!10^{-16}$ & [-0.1888,\,-0.1277] \\
FITS               & 0.061 & -0.2398 & -0.1988 & $2.99\!\times\!10^{-28}$ & [-0.2730,\,-0.2076] & 0.094 & -0.2138 & -0.1651 & $5.35\!\times\!10^{-25}$ & [-0.2496,\,-0.1779] \\
GDN                & 0.156 & -0.1647 & -0.1067 & $1.01\!\times\!10^{-20}$ & [-0.1964,\,-0.1345] & 0.189 & -0.0870 & -0.0667 & $4.46\!\times\!10^{-14}$ & [-0.1108,\,-0.0643] \\
HBOS               & 0.194 & -0.2759 & -0.2588 & $2.24\!\times\!10^{-19}$ & [-0.3224,\,-0.2293] & 0.128 & -0.1916 & -0.1984 & $2.23\!\times\!10^{-20}$ & [-0.2217,\,-0.1611] \\
IForest            & 0.217 & -0.2274 & -0.1814 & $5.74\!\times\!10^{-15}$ & [-0.2780,\,-0.1776] & 0.200 & -0.1207 & -0.1122 & $5.24\!\times\!10^{-15}$ & [-0.1503,\,-0.0908] \\
KMeansAD           & 0.239 & -0.1853 & -0.1482 & $4.22\!\times\!10^{-12}$ & [-0.2335,\,-0.1336] & 0.261 & -0.1337 & -0.1174 & $3.58\!\times\!10^{-14}$ & [-0.1646,\,-0.1023] \\
KNN                & 0.128 & -0.3043 & -0.2596 & $1.09\!\times\!10^{-25}$ & [-0.3452,\,-0.2632] & 0.039 & -0.3257 & -0.3305 & $1.04\!\times\!10^{-29}$ & [-0.3554,\,-0.2940] \\
LOF                & 0.133 & -0.3406 & -0.2886 & $2.50\!\times\!10^{-25}$ & [-0.3881,\,-0.2962] & 0.039 & -0.2906 & -0.3184 & $7.00\!\times\!10^{-30}$ & [-0.3126,\,-0.2674] \\
LSTMAD             & 0.144 & -0.1893 & -0.1256 & $9.95\!\times\!10^{-21}$ & [-0.2249,\,-0.1550] & 0.139 & -0.2274 & -0.1489 & $3.90\!\times\!10^{-22}$ & [-0.2673,\,-0.1873] \\
M2N2               & 0.239 & -0.1181 & -0.0955 & $3.82\!\times\!10^{-10}$ & [-0.1695,\,-0.0663] & 0.217 & -0.0851 & -0.0866 & $2.90\!\times\!10^{-10}$ & [-0.1162,\,-0.0542] \\
MAVAE              & 0.322 & -0.1380 & -0.0390 & $2.07\!\times\!10^{-9}$  & [-0.1802,\,-0.0936] & 0.317 & -0.1273 & -0.0279 & $6.14\!\times\!10^{-10}$ & [-0.1621,\,-0.0932] \\
OFA                & 0.217 & -0.1368 & -0.0907 & $1.09\!\times\!10^{-15}$ & [-0.1715,\,-0.1029] & 0.206 & -0.1855 & -0.1181 & $1.35\!\times\!10^{-16}$ & [-0.2228,\,-0.1473] \\
OmniAnomaly         & 0.328 & -0.0649 & -0.0190 & $2.76\!\times\!10^{-5}$  & [-0.0961,\,-0.0354] & 0.372 & -0.0806 & -0.0043 & $1.94\!\times\!10^{-5}$  & [-0.1097,\,-0.0518] \\
PCA                & 0.161 & -0.1949 & -0.1406 & $1.09\!\times\!10^{-15}$ &[-0.2428,\,-0.1483] & 0.178 & -0.1488 & -0.0990 & $2.65\!\times\!10^{-19}$ &[-0.1793,\,-0.1211] \\
RobustPCA          & 0.150 & -0.1989 & -0.2100 & $1.05\!\times\!10^{-15}$ & [-0.2548,\,-0.1407] & 0.106 & -0.2354 & -0.2480 & $3.91\!\times\!10^{-19}$ & [-0.2782,\,-0.1953] \\
SISVAE             & 0.272 & -0.1143 & -0.0522 & $1.73\!\times\!10^{-11}$ & [-0.1511,\,-0.0781] & 0.322 & -0.0659 & -0.0157 & $9.74\!\times\!10^{-7}$  & [-0.0917,\,-0.0403] \\
TFTResidual        & 0.133 & -0.1868 & -0.1252 & $4.79\!\times\!10^{-21}$ & [-0.2212,\,-0.1509] & 0.206 & -0.1152 & -0.0852 & $1.54\!\times\!10^{-14}$ & [-0.1429,\,-0.0886] \\
TimesNet           & 0.117 & -0.2355 & -0.1916 & $5.03\!\times\!10^{-27}$ & [-0.2693,\,-0.2038] & 0.133 & -0.2068 & -0.1294 & $1.99\!\times\!10^{-22}$ & [-0.2433,\,-0.1702] \\
TranAD             & 0.194 & -0.1793 & -0.1080 & $1.24\!\times\!10^{-18}$ & [-0.2153,\,-0.1433] & 0.161 & -0.1498 & -0.1024 & $5.06\!\times\!10^{-19}$ & [-0.1791,\,-0.1197] \\
USAD               & 0.294 & -0.0744 & -0.0275 & $2.14\!\times\!10^{-7}$  & [-0.1044,\,-0.0451] & 0.294 & -0.0867 & -0.0084 & $2.33\!\times\!10^{-8}$  & [-0.1159,\,-0.0580] \\
VASP               & 0.289 & -0.0978 & -0.0449 & $1.83\!\times\!10^{-10}$ & [-0.1286,\,-0.0676] & 0.333 & -0.0626 & -0.0158 & $1.13\!\times\!10^{-6}$  & [-0.0857,\,-0.0400] \\
VSVAE              & 0.333 & -0.1465 & -0.0592 & $8.43\!\times\!10^{-9}$  & [-0.1927,\,-0.1017] & 0.378 & -0.1159 & -0.0281 & $1.17\!\times\!10^{-6}$  & [-0.1525,\,-0.0778] \\
WVAE               & 0.322 & -0.0829 & -0.0381 & $5.57\!\times\!10^{-7}$  & [-0.1182,\,-0.0451] & 0.372 & -0.0775 & -0.0103 & $8.81\!\times\!10^{-5}$  & [-0.1086,\,-0.0479] \\
\bottomrule
\end{tabular}%
}
\end{table}

\begin{table}[htbp]
\centering
\caption{Continuation of Table~\ref{tab:tsbad_pairwise_vs_axonad}: VUS-PR, R-based-F1, and Affiliation-F1 vs AxonAD. Missing metrics in your export (VUS-ROC, PA-F1, Event-F1, Range-F1) are not shown.}
\label{tab:tsbad_pairwise_vs_axonad_2}
\resizebox{\textwidth}{!}{%
\begin{tabular}{@{}lccccccccccccccc@{}}
\toprule
\multirow{2}{*}{\textbf{Model}}
& \multicolumn{5}{c}{\textbf{VUS-PR (vs AxonAD)}}
& \multicolumn{5}{c}{\textbf{R-based-F1 (vs AxonAD)}}
& \multicolumn{5}{c}{\textbf{Affiliation-F (vs AxonAD)}} \\
\cmidrule(l){2-6}\cmidrule(l){7-11}\cmidrule(l){12-16}
& \textbf{WR} & $\Delta$ mean & $\Delta$ med & $p$ & CI$_{95}$
& \textbf{WR} & $\Delta$ mean & $\Delta$ med & $p$ & CI$_{95}$
& \textbf{WR} & $\Delta$ mean & $\Delta$ med & $p$ & CI$_{95}$ \\
\midrule
AnomalyTransformer & 0.039 & -0.3784 & -0.2998 & $1.65\!\times\!10^{-30}$ & [-0.4217,\,-0.3367]
& 0.067 & -0.3325 & -0.3168 & $2.57\!\times\!10^{-29}$ & [-0.3678,\,-0.2964]
& 0.200 & -0.1228 & -0.0718 & $3.64\!\times\!10^{-18}$ & [-0.1478,\,-0.0993] \\
AutoEncoder        & 0.172 & -0.1987 & -0.1348 & $1.88\!\times\!10^{-16}$ & [-0.2422,\,-0.1567]
& 0.189 & -0.1882 & -0.2248 & $2.51\!\times\!10^{-13}$ & [-0.2396,\,-0.1352]
& 0.333 & -0.0596 & -0.0207 & $6.96\!\times\!10^{-7}$  & [-0.0831,\,-0.0354] \\
CBLOF              & 0.161 & -0.2334 & -0.1601 & $1.49\!\times\!10^{-19}$ & [-0.2776,\,-0.1924]
& 0.200 & -0.1692 & -0.2151 & $1.54\!\times\!10^{-12}$ & [-0.2177,\,-0.1185]
& 0.361 & -0.0487 & -0.0081 & $8.71\!\times\!10^{-5}$  & [-0.0752,\,-0.0225] \\
CNN                & 0.278 & -0.1408 & -0.0769 & $9.59\!\times\!10^{-12}$ & [-0.1805,\,-0.1039]
& 0.217 & -0.1694 & -0.1844 & $3.38\!\times\!10^{-13}$ & [-0.2190,\,-0.1180]
& 0.483 &  0.0062 & -0.0002 & \textbf{0.676} & [-0.0127,\,0.0264] \\
COPOD              & 0.150 & -0.2900 & -0.2262 & $3.03\!\times\!10^{-22}$ & [-0.3360,\,-0.2450]
& 0.183 & -0.2287 & -0.2535 & $6.43\!\times\!10^{-15}$ & [-0.2815,\,-0.1741]
& 0.333 & -0.0609 & -0.0538 & $2.21\!\times\!10^{-7}$  & [-0.0853,\,-0.0363] \\
Donut              & 0.100 & -0.2312 & -0.1677 & $2.85\!\times\!10^{-23}$ & [-0.2700,\,-0.1925]
& 0.100 & -0.2913 & -0.2873 & $7.64\!\times\!10^{-27}$ & [-0.3273,\,-0.2543]
& 0.261 & -0.0913 & -0.0147 & $1.59\!\times\!10^{-13}$ & [-0.1181,\,-0.0681] \\
EIF                & 0.161 & -0.2836 & -0.1920 & $2.10\!\times\!10^{-23}$ & [-0.3275,\,-0.2431]
& 0.189 & -0.2125 & -0.2535 & $3.91\!\times\!10^{-14}$ & [-0.2640,\,-0.1614]
& 0.367 & -0.0477 & -0.0163 & $1.59\!\times\!10^{-5}$  & [-0.0704,\,-0.0255] \\
FITS               & 0.100 & -0.2265 & -0.1607 & $1.92\!\times\!10^{-26}$ & [-0.2615,\,-0.1927]
& 0.083 & -0.2898 & -0.2566 & $9.49\!\times\!10^{-27}$ & [-0.3260,\,-0.2536]
& 0.306 & -0.0445 & -0.0226 & $1.56\!\times\!10^{-7}$  & [-0.0606,\,-0.0286] \\
GDN                & 0.161 & -0.1611 & -0.0866 & $1.87\!\times\!10^{-19}$ & [-0.1936,\,-0.1294]
& 0.111 & -0.2625 & -0.2455 & $1.79\!\times\!10^{-25}$ & [-0.2969,\,-0.2273]
& 0.372 & -0.0145 & -0.0024 & \textbf{0.0621} & [-0.0308,\,0.0024] \\
HBOS               & 0.128 & -0.3030 & -0.2395 & $9.54\!\times\!10^{-25}$ & [-0.3484,\,-0.2584]
& 0.189 & -0.2271 & -0.2438 & $7.61\!\times\!10^{-15}$ & [-0.2803,\,-0.1770]
& 0.333 & -0.0637 & -0.0600 & $3.80\!\times\!10^{-8}$  & [-0.0855,\,-0.0419] \\
IForest            & 0.189 & -0.2399 & -0.2016 & $8.52\!\times\!10^{-17}$ & [-0.2862,\,-0.1896]
& 0.217 & -0.2275 & -0.2268 & $9.05\!\times\!10^{-18}$ & [-0.2689,\,-0.1849]
& 0.339 & -0.0590 & -0.0399 & $4.68\!\times\!10^{-9}$  & [-0.0767,\,-0.0401] \\
KMeansAD           & 0.239 & -0.1970 & -0.1592 & $2.18\!\times\!10^{-14}$ & [-0.2403,\,-0.1514]
& 0.289 & -0.1449 & -0.1326 & $1.06\!\times\!10^{-8}$  & [-0.1866,\,-0.0998]
& 0.422 & -0.0407 & -0.0138 & $1.06\!\times\!10^{-4}$  & [-0.0599,\,-0.0196] \\
KNN                & 0.061 & -0.3168 & -0.2638 & $1.46\!\times\!10^{-28}$ & [-0.3565,\,-0.2764]
& 0.167 & -0.2660 & -0.2216 & $6.33\!\times\!10^{-23}$ & [-0.3040,\,-0.2264]
& 0.300 & -0.0695 & -0.0435 & $2.44\!\times\!10^{-9}$  & [-0.0909,\,-0.0484] \\
LOF                & 0.067 & -0.3548 & -0.2799 & $2.79\!\times\!10^{-29}$ & [-0.3986,\,-0.3115]
& 0.133 & -0.3220 & -0.2865 & $1.54\!\times\!10^{-25}$ & [-0.3641,\,-0.2781]
& 0.279 & -0.0953 & -0.0677 & $9.44\!\times\!10^{-12}$ & [-0.1179,\,-0.0716] \\
LSTMAD             & 0.144 & -0.2482 & -0.1504 & $8.26\!\times\!10^{-22}$ & [-0.2914,\,-0.2049]
& 0.094 & -0.2726 & -0.2264 & $1.95\!\times\!10^{-27}$ & [-0.3066,\,-0.2361]
& 0.344 & -0.1594 & -0.0079 & $6.26\!\times\!10^{-8}$  & [-0.2043,\,-0.1159] \\
M2N2               & 0.228 & -0.1702 & -0.0953 & $5.15\!\times\!10^{-14}$ & [-0.2105,\,-0.1322]
& 0.206 & -0.1889 & -0.2161 & $3.35\!\times\!10^{-14}$ & [-0.2381,\,-0.1375]
& 0.439 & -0.0004 & -0.0011 & \textbf{0.376} & [-0.0183,\,0.0175] \\
MAVAE              & 0.300 & -0.1425 & -0.0495 & $2.12\!\times\!10^{-10}$ & [-0.1839,\,-0.0997]
& 0.228 & -0.1455 & -0.0838 & $1.20\!\times\!10^{-15}$ & [-0.1805,\,-0.1105]
& 0.322 & -0.0484 & -0.0026 & $4.08\!\times\!10^{-8}$  & [-0.0648,\,-0.0310] \\
OFA                & 0.222 & -0.1260 & -0.0797 & $3.25\!\times\!10^{-14}$ & [-0.1600,\,-0.0930]
& 0.283 & -0.1111 & -0.0820 & $4.37\!\times\!10^{-11}$ & [-0.1419,\,-0.0796]
& 0.367 & -0.0275 & -0.0035 & $7.76\!\times\!10^{-4}$  & [-0.0429,\,-0.0124] \\
OmniAnomaly        & 0.328 & -0.0697 & -0.0172 & $6.67\!\times\!10^{-6}$  & [-0.0998,\,-0.0396]
& 0.361 & -0.0393 & -0.0153 & $3.44\!\times\!10^{-3}$  & [-0.0691,\,-0.0107]
& 0.422 & -0.0186 & -0.0002 & \textbf{0.0852} & [-0.0358,\,-0.0010] \\
PCA                & 0.161 & -0.2167 & -0.1472 & $2.20\!\times\!10^{-18}$ & [-0.2611,\,-0.1722]
& 0.267 & -0.1456 & -0.0754 & $3.45\!\times\!10^{-11}$ &[-0.1832,\,-0.1076]
& 0.317 & -0.0712 & -0.0198 & $1.59\!\times\!10^{-10}$ &[-0.0891,\,-0.0532] \\
RobustPCA          & 0.128 & -0.2550 & -0.2011 & $4.81\!\times\!10^{-22}$ & [-0.3000,\,-0.2135]
& 0.267 & -0.1393 & -0.1996 & $1.13\!\times\!10^{-8}$  & [-0.1946,\,-0.0826]
& 0.294 & -0.0709 & -0.0578 & $1.68\!\times\!10^{-8}$  & [-0.0964,\,-0.0461] \\
SISVAE             & 0.233 & -0.1215 & -0.0660 & $4.74\!\times\!10^{-13}$ & [-0.1569,\,-0.0870]
& 0.289 & -0.1016 & -0.0493 & $1.16\!\times\!10^{-9}$  & [-0.1368,\,-0.0688]
& 0.383 & -0.0360 & -0.0005 & $1.56\!\times\!10^{-4}$  & [-0.0529,\,-0.0188] \\
TFTResidual        & 0.150 & -0.1857 & -0.1180 & $5.43\!\times\!10^{-20}$ & [-0.2228,\,-0.1480]
& 0.139 & -0.2636 & -0.2759 & $3.74\!\times\!10^{-25}$ & [-0.2979,\,-0.2285]
& 0.400 & -0.0136 & -0.0024 & \textbf{0.0563} & [-0.0307,\,0.0041] \\
TimesNet           & 0.100 & -0.2222 & -0.1716 & $3.55\!\times\!10^{-26}$ & [-0.2572,\,-0.1889]
& 0.089 & -0.2952 & -0.2675 & $1.66\!\times\!10^{-27}$ & [-0.3304,\,-0.2586]
& 0.306 & -0.0390 & -0.0161 & $8.00\!\times\!10^{-7}$  & [-0.0547,\,-0.0224] \\
TranAD             & 0.233 & -0.1848 & -0.0947 & $1.27\!\times\!10^{-16}$ & [-0.2241,\,-0.1479]
& 0.128 & -0.2533 & -0.2310 & $5.43\!\times\!10^{-26}$ & [-0.2868,\,-0.2187]
& 0.383 & -0.0337 & -0.0035 & $6.61\!\times\!10^{-4}$  & [-0.0499,\,-0.0167] \\
USAD               & 0.289 & -0.0812 & -0.0253 & $1.98\!\times\!10^{-8}$  & [-0.1111,\,-0.0518]
& 0.350 & -0.0490 & -0.0214 & $1.16\!\times\!10^{-3}$  & [-0.0774,\,-0.0200]
& 0.411 & -0.0229 & -0.0012 & $2.56\!\times\!10^{-2}$  & [-0.0397,\,-0.0050] \\
VASP               & 0.256 & -0.0927 & -0.0427 & $4.00\!\times\!10^{-10}$ & [-0.1235,\,-0.0622]
& 0.350 & -0.0705 & -0.0641 & $6.53\!\times\!10^{-6}$  & [-0.0999,\,-0.0422]
& 0.444 & -0.0115 & -0.0012 & \textbf{0.139} & [-0.0275,\,0.0058] \\
VSVAE              & 0.339 & -0.1512 & -0.0672 & $4.18\!\times\!10^{-9}$  & [-0.1971,\,-0.1069]
& 0.339 & -0.0967 & -0.0449 & $1.17\!\times\!10^{-6}$  & [-0.1337,\,-0.0611]
& 0.472 & -0.0189 & -0.0000 & \textbf{0.120} & [-0.0348,\,-0.0029] \\
WVAE               & 0.300 & -0.0802 & -0.0363 & $8.24\!\times\!10^{-7}$  & [-0.1156,\,-0.0426]
& 0.250 & -0.1062 & -0.0697 & $6.29\!\times\!10^{-11}$ & [-0.1386,\,-0.0721]
& 0.422 & -0.0216 & -0.0004 & $4.16\!\times\!10^{-3}$  & [-0.0386,\,-0.0039] \\
\bottomrule
\end{tabular}%
}
\end{table}

\end{document}